\documentclass[10pt,twocolumn,letterpaper]{article}

\usepackage[pagenumbers]{cvpr} 

\usepackage{amsthm}
\usepackage{graphicx}
\usepackage{animate}
\newcommand{\Var}{\textup{Var}}
\newcommand{\Cov}{\textup{Cov}}
\newcommand{\bE}{\mathbb{E}}
\newtheorem{proposition}{Proposition}
\newtheorem{example}{Example}

\definecolor{cvprblue}{rgb}{0.21,0.49,0.74}
\usepackage[pagebackref,breaklinks,colorlinks,allcolors=cvprblue]{hyperref}
\usepackage{mathtools}
\usepackage{color,colortbl}
\usepackage{arydshln}
\usepackage{comment}
\usepackage{algorithm}
\usepackage{algpseudocode}
\usepackage{listings}
\usepackage{xcolor} 
\lstdefinestyle{pythonstyle}{
    language=Python,
    basicstyle=\ttfamily\scriptsize, 
    keywordstyle=\bfseries\color[rgb]{0.1, 0.5, 0.1},
    morekeywords=[2]{in, and},
    keywordstyle=[2]\bfseries\color[rgb]{0.8, 0.2, 0.5},
    morekeywords=[3]{warp_noise},
    keywordstyle=[3]\color[rgb]{0.1, 0.1, 0.8},
    commentstyle=\color[rgb]{0.2, 0.6, 0.2},
    stringstyle=\color[rgb]{0.0, 0.4, 0.2},
    numbers=left,
    numberstyle=\tiny\color{gray},
    stepnumber=1,
    numbersep=5pt,
    backgroundcolor=\color{white},
    frame=tb,
    framerule=0.5pt,
    rulecolor=\color{black},
    framesep=2mm,
    xleftmargin=2mm,
    xrightmargin=2mm,
    breaklines=true,
    captionpos=b,
    showstringspaces=false
}


\def\paperID{17653} 
\def\confName{CVPR}
\def\confYear{2025}

\newcommand{\ning}[1]{{\color{black}#1}}

\newcommand{\RYAN}[1]{#1}

\newcommand{\method}{Go-with-the-Flow\xspace}
\newcommand{\cutndrag}{cut-and-drag}

\newcommand{\deglevel}{\mathbf{\gamma}}
\newcommand{\Q}{\mathbf{Q}}
\newcommand{\expansion}{\textit{expansion}}
\newcommand{\contraction}{\textit{contraction}}

\newcommand*{\affaddr}[1]{#1}
\newcommand*{\affmark}[1][*]{\textsuperscript{#1}}
\newcommand{\customfootnotetext}[2]{{
  \renewcommand{\thefootnote}{#1}
  \footnotetext[0]{#2}}}

\title{\textbf{\method}:\\Motion-Controllable Video Diffusion Models Using Real-Time Warped Noise}

\author{Ryan Burgert\affmark[1,3]\hspace{0.5cm}
Yuancheng Xu\affmark[1,4]\hspace{0.5cm}
Wenqi Xian\affmark[1]\hspace{0.5cm}
Oliver Pilarski\affmark[1]\hspace{0.5cm}
Pascal Clausen\affmark[1]\\
Mingming He\affmark[1]\hspace{0.5cm}
Li Ma\affmark[1]\hspace{0.5cm}
Yitong Deng\affmark[2,5]\hspace{0.5cm}
Lingxiao Li\affmark[2]\hspace{0.5cm}
Mohsen Mousavi\affmark[1]\hspace{0.5cm}
Michael Ryoo\affmark[3]\\
Paul Debevec\affmark[1]\hspace{0.5cm}
Ning Yu\affmark[1]\textsuperscript{$\dagger$}\\
\\
\affaddr{\affmark[1]Netflix Eyeline Studios}\hspace{0.5cm}
\affaddr{\affmark[2]Netflix}\hspace{0.5cm}
\affaddr{\affmark[3]Stony Brook University}\\
\affaddr{\affmark[4]University of Maryland}\hspace{0.5cm}
\affaddr{\affmark[5]Stanford University}\\
\tt\small \{ryan.burgert,yuancheng.xu,wenqi.xian,oliver.pilarski,pascal.clausen,\\
\tt\small mingming.he,li.ma,mohsen.mousavi,debevec,ning.yu\}@scanlinevfx.com \\
\tt\small lingxiaol@netflix.com\hspace{0.5cm}
\tt\small \{rburgert,mryoo\}@cs.stonybrook.edu\\
\tt\small ycxu@umd.edu\hspace{0.5cm}
\tt\small yitongd@stanford.edu
\\
\href{https://eyeline-labs.github.io/Go-with-the-Flow/}{https://eyeline-labs.github.io/Go-with-the-Flow/}
}

\begin{document}

\twocolumn[{%
\renewcommand\twocolumn[1][]{#1}%
\vspace{-40pt}
\maketitle
\vspace{-35pt}
\begin{center}
    \centering
    \captionsetup{type=figure}
    \includegraphics[width=\linewidth]{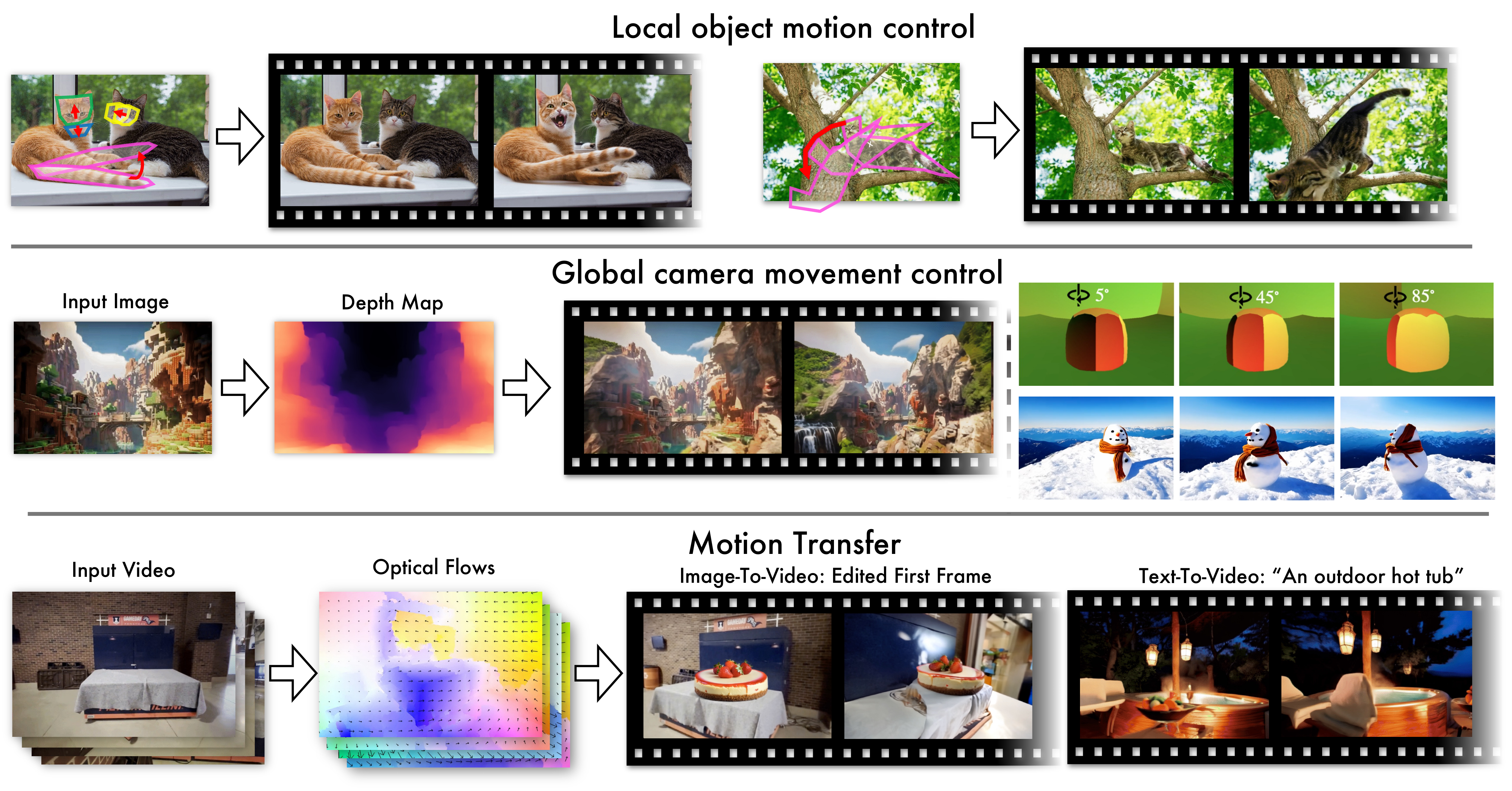}
    \captionof{figure}{\method presents a simple, robust, and easy-to-implement method for motion-controllable video diffusion models based on optical flow and noise warping. It only requires fine-tuning video diffusion models as a black box using warped noise patterns. Leveraging our models, we can (1) control the motion of individual objects or parts of those objects, (2) direct the camera movement by providing global flow fields corresponding to the desired movements, and (3) transfer the motion from input videos to target contexts.}
    \label{fig:teaser}
\end{center}
}]

\customfootnotetext{$\dagger$}{Project lead}

\begin{abstract}

Generative modeling aims to transform random noise into structured outputs. In this work, we enhance video diffusion models by allowing motion control via structured latent noise sampling. This is achieved by just a change in data: we pre-process training videos to yield structured noise. Consequently, our method is agnostic to diffusion model design, requiring no changes to model architectures or training pipelines. Specifically, we propose a novel noise warping algorithm, fast enough to run in real time, that replaces random temporal Gaussianity with correlated warped noise derived from optical flow fields, while preserving the spatial Gaussianity. The efficiency of our algorithm enables us to fine-tune modern video diffusion base models using warped noise with minimal overhead, and provide a one-stop solution for a wide range of user-friendly motion control: local object motion control, global camera movement control, and motion transfer. The harmonization between temporal coherence and spatial Gaussianity in our warped noise leads to effective motion control while maintaining per-frame pixel quality. Extensive experiments and user studies demonstrate the advantages of our method, making it a robust and scalable approach for controlling motion in video diffusion models. Video results are available on our \href{https://eyeline-research.github.io/Go-with-the-Flow/}{webpage}; source code and model checkpoints are available on \href{https://github.com/Eyeline-Research/Go-with-the-Flow}{GitHub}.

\end{abstract}
\section{Introduction}
\label{sec:introduction}

\begin{figure*}
    \centering
    \includegraphics[width=\linewidth]{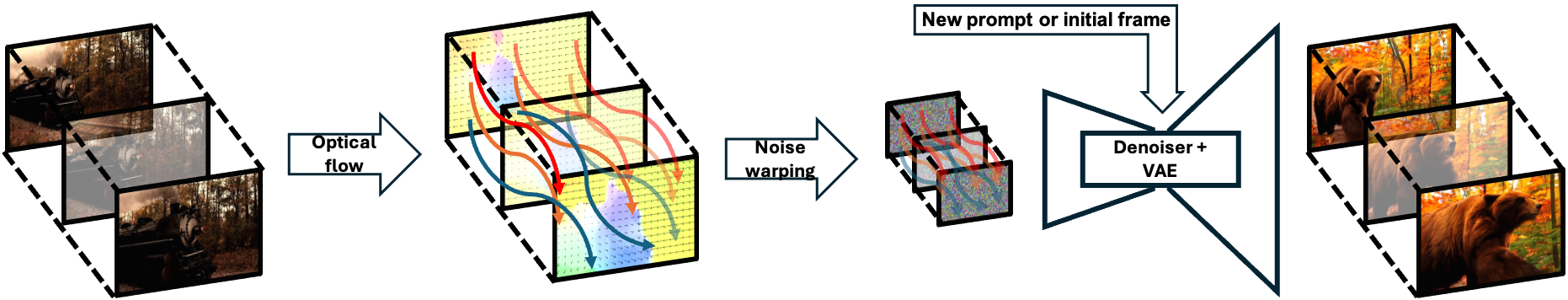}
    \caption{Our method consists of three components: flow field extraction, real-time noise warping, and diffusion model fine-tuning/inference. During fine-tuning, we use the original captions of video samples. At inference, our method enables adaptation of reference motion to various prompts and/or initial frames, offering creativity and diversity in generation.
    }
    \label{fig:framework_diagram}
\end{figure*}

\begin{quote}
    “\textit{We adore chaos because we love to produce order.}” --- M. C. Escher, Dutch artist
\end{quote}

The essence of generative modeling lies in producing order from chaos, learning to transform random noise from the latent space into structured outputs that align with the distribution of training data. In this paper, we propose a novel approach to enhance generative model learning by proactively introducing partial order into the chaos of latent space sampling.

Our work is motivated by the remarkable progress in video diffusion generative models~\cite{guo2024animatediff,blattmann2023align,guo2024sparsectrl,blattmann2023stable,brooks2024video,yang2024cogvideox} and the equally significant challenges they face in terms of controllability beyond text and image guidance. Fine-grained interactive control over motion dynamics remains an under-explored area due to the intricate spatiotemporal correlations among video frames. The complexity of modern video diffusion architectures~\cite{brooks2024video,yang2024cogvideox}, which leverage 3D autoencoders~\cite{yu2024language} and spatiotemporal tokenizers~\cite{ma2024latte}, further complicates efforts to adapt models for effective motion control. The optimal format for defining and disentangling motion control from other guidance remains an open question.

Within the domain of motion-controllable video diffusion models, current applications typically fall into three categories: (1) local object motion control, represented by object bounding boxes or masks with motion trajectories~\cite{jain2024peekaboo,yang2024direct,wang2024motionctrl,shi2024motion,wu2024draganything,namekata2024sg,qiu2024freetraj,geng2024motion}; (2) global camera movement control, parameterized by camera poses and trajectories~\cite{yang2024direct,he2024cameractrl,kuang2024collaborative,wang2024motionctrl,xu2024camco,wu2024cat4d} or categorized by common directional patterns such as panning and tilting~\cite{guo2024animatediff,yang2024direct}; and (3) motion transfer from reference videos to target contexts specified by prompts or initial frames~\cite{wang2024videocomposer,yatim2024space,geyer2023tokenflow,yin2024scalable,ku2024anyv2v,ling2024motionclone,mou2024revideo,aira2024motioncraft}. However, these approaches share three key limitations: (1) they often necessitate complex modifications to the base model design such as guidance attention~\cite{qiu2024freetraj}, limiting compatibility with modern full-attention architectures involving spatiotemporal tokens~\cite{yang2024cogvideox}; (2) they are constrained to specific applications, requiring detailed parameterized motion signals, such as camera parameters, which are challenging to acquire or estimate accurately~\cite{zhang2024monst3r}, thus restricting generalizability across diverse scenarios; and (3) they are over-rigid to motion control at the cost of spatiotemporal visual quality.

To address these limitations, we propose a novel and straightforward method to incorporate motion control as a structured component within the chaos of video diffusion's latent space. We achieve this by correlating the temporal distribution of latent noise. Specifically, starting with a 2D Gaussian noise slice, we temporally concatenate it with warped noise slices, given the optical flow field~\cite{teed2020raft} extracted from a training video sample. Fig.~\ref{fig:framework_diagram} illustrates the diagram of our method. Our approach requires only a change in data: we pre-process training videos to yield warped noise and then fine-tune a video diffusion model. As it occurs solely during noise sampling, our method is agnostic to diffusion model design, requiring no modifications to model architectures or training pipelines. Surprisingly, removing temporal Gaussianity from the noise distribution does not deteriorate model fine-tuning. Instead, it can be quickly adapted after fine-tuning because temporal structure in the chaos of latent space facilitates generative learning and enables motion correspondence. Temporal coherence occurring in the latent space also harmonizes motion control with per-frame pixel quality by inheriting the high-quality prior from the base model.

It is worth noting that video diffusion fine-tuning relies on efficient noise warping algorithms that introduce minimal overhead during data pre-processing and noise sampling. The existing noise warping algorithm, \textit{How I Warped Your Noise} (HIWYN)~\cite{chang2024warped}, that maintains spatial Gaussianity and enables temporal flow warping, however, suffers from the quadratic computation costs w.r.t. frame count, making it \ning{much slower than in real time and therefore} impractical for large-scale video diffusion model training. To address this, we propose a novel noise warping algorithm \ning{that runs fast enough in real time}. Rather than warping each frame through a chain of operations from the initial frame, our algorithm iteratively warps noise between consecutive frames. This is achieved by carefully tracking the noise and the flow density along a forward and a backward flow at the pixel level, accounting for both expansion and contraction dynamics, supplemented with conditional white noise sampling from HIWYN~\citet{chang2024warped} to preserve Gaussianity. \cref{alg:main} provides further details. We validate the spatial Gaussianity and time complexity of our noise-warping algorithm and apply it to training-free image diffusion models for quantitative and qualitative assessments of controllability and temporal consistency.

During video diffusion inference, our method offers a one-stop solution for diverse motion control applications by adapting noise warping based on motion type. (1) For local object motion, we interactively transform noise elements within object masks given users' dragging signals. (2) For global camera movement control, we reuse the optical flows from reference videos to warp input noise, and regenerate videos conditioned on different texts or initial frames. (3) For arbitrary motion transfer, the motion representations are not limited to optical flows~\cite{teed2020raft}, but also include flows from 3D rendering engines~\cite{villar2021learning}, depth warping~\cite{yu2024wonderjourney}, etc. We validate the effectiveness of our solution across various video generation tasks, demonstrating its ability to preserve consistent motion across different contexts or render distinct motions for the same context. Extensive experiments and user studies indicate the advantages of our solution in pixel quality, motion control, text alignment, temporal consistency, and user preference.

In summary, our contributions include:

(1) A novel and simple one-stop solution for motion-controllable video diffusion models, integrating motion control as a flow field for noise warping in latent space sampling, plug-and-play for any video diffusion base models as a black box, and compatible with any other types of controls.

(2) An efficient noise warping algorithm that maintains spatial Gaussianity and follows temporal motion flows across frames, facilitating motion-controllable video diffusion model fine-tuning with minimal overhead.

(3) Comprehensive experiments and user studies demonstrating the overall advantageous pixel quality, controllability, temporal consistency, and subjective preference of our method on diverse motion control applications, including but not limited to: local object motion control, motion transfer to new contexts, and reference-based global camera movement control.

\section{Related work}
\label{sec:related_work}

\subsection{Image and video diffusion models}

With the theoretical establishments of diffusion models~\cite{song2021score,ho2020denoising,song2021denoising,karras2022elucidating} and their practical advancements~\cite{nichol2021glide,ho2022classifier}, and when sophisticated text encoders~\cite{radford2021learning} and language models~\cite{raffel2020exploring} meet diffusion models, great breakthroughs in text-to-image generation~\cite{rombach2022high,podell2023sdxl,stabilityai2023deepfloyd} have revolutionized how we digitize and create visual worlds. Building upon these, image-to-image diffusion models~\cite{brooks2023instructpix2pix,zhang2023adding,ke2024repurposing} enable image editing applications like stylization~\cite{meng2022sdedit}, relighting~\cite{he2024diffrelight}, and super-resolution~\cite{yue2024resshift,stabilityai2023deepfloyd}, expanding creativity in recreating or enhancing visual worlds.

A natural extension of image generation use cases is to cover the temporal dimension for video generation. The most cost-efficient way is to reuse the well-trained image diffusion model weights. Directly querying the above image diffusion models using random noise to generate videos frame-by-frame often struggles with temporal inconsistency, flickering, or semantic drifting. Noise warping, HIWYN~\cite{chang2024warped}, as a method for creating a sequence of temporally-correlated latent noise from optical flow while claiming spatial Gaussianity preservation, yields temporally consistent motion patterns after querying image diffusion models without further fine-tuning. To overcome its defective spatial Gaussianity preservation and undesired time complexity, we propose a novel warped noise sampling algorithm that guarantees spatial Gaussianity and runs fast enough in real time. We validate its efficacy by applying it to the training-free image diffusion models like DifFRelight~\cite{he2024diffrelight} for video relighting and DeepFloyd IF~\cite{stabilityai2023deepfloyd} for video super-resolution.

Video diffusion model training is a more costly yet more effective way for video generation~\cite{brooks2024video,chen2023videocrafter1,blattmann2023align,guo2024animatediff,blattmann2023stable,qing2024hierarchical,xing2024dynamicrafter,yang2024cogvideox}. AnimateDiff~\cite{guo2024animatediff} upgrades pre-trained image diffusion models by fine-tuning temporal attention layers on large-scale video datasets. CogVideoX~\cite{yang2024cogvideox}, a state-of-the-art open-source video diffusion model, combines spatial and temporal dimensions by encoding/decoding videos via 3D causal VAE~\cite{yu2024language} and diffusing/denoising spatiotemporal tokens via diffusion transformers~\cite{peebles2023scalable}. We use CogVideoX~\cite{yang2024cogvideox} as a base model and incorporate our warped noise sampling for motion-controllable fine-tuning. We also fine-tune on AnimateDiff~\cite{guo2024animatediff} to show our method is model-agnostic.

\subsection{Motion controllable video generation}

Beyond text~\cite{guo2024animatediff,yang2024cogvideox} and image controls~\cite{guo2024sparsectrl,xing2024dynamicrafter,zhou2024storydiffusion} for video diffusion models, motion control makes video generation more interactive, dynamically targeted, and spatiotemporally fine-grained. Current approaches to motion control follow three main paradigms:

Firstly, \textit{local object motion control} is represented by object bounding boxes or masks with motion trajectories~\cite{jain2024peekaboo,yang2024direct,wang2024motionctrl,shi2024motion,wu2024draganything,namekata2024sg,qiu2024freetraj,geng2024motion}. DragAnything~\cite{wu2024draganything} allows precise object motion manipulation in images without retraining, while SG-I2V~\cite{namekata2024sg} generates realistic, continuous video from single images using self-guided motion trajectories. These serve as recent baselines for local object motion control. Our method is plug-and-play, treating diffusion models as a black box while using synthetic flows to mimic and densify object trajectories at the pixel level.

Secondly, \textit{global camera movement control} is parameterized by camera poses and trajectories~\cite{yang2024direct,he2024cameractrl,kuang2024collaborative,wang2024motionctrl,xu2024camco,wu2024cat4d} or categorized by common directional patterns like panning and tilting~\cite{guo2024animatediff,yang2024direct}. These methods introduce additional modules that accept camera parameters, trained in a supervised manner. Other approaches~\cite{yu2024viewcrafter,hou2024training} leverage rendering priors as input for camera control. Approaches like ReCapture~\cite{zhang2024recapture} enable reconfiguration of camera trajectories in given videos. Our method bypasses the need for extensive camera parameter collection, and directly generalizes new camera movements from reference videos at inference.

Lastly, \textit{motion transfer} happens from reference videos to target contexts~\cite{wang2024videocomposer,yatim2024space,geyer2023tokenflow,yin2024scalable,ku2024anyv2v,ling2024motionclone,mou2024revideo,aira2024motioncraft}. DiffusionMotionTransfer~\cite{yatim2024space} introduces a loss that maintains scene layout and motion fidelity in target videos, while MotionClone~\cite{ling2024motionclone} uses temporal attention as motion representation, streamlining motion transfer. Using them as motion transfer baselines, we demonstrate our model's flexibility in combining reference geometries with target text guidance.

\section{Method}
\label{sec:method}

\method is comprised of two separate parts: our noise warping algorithm and video diffusion fine-tuning. The noise warping algorithm operates independently from the diffusion model training process: we use the noise patterns it produces to train the diffusion model. Our motion control is based \textit{entirely} on noise initializations, introducing no extra parameters to the video diffusion model.

Inspired by the existing noise warping algorithm HIWYN~\cite{chang2024warped}, which introduced noise warping for image diffusion models, we introduce a new use case for the warped noise: we use it as a form of \textit{motion conditioning} for video generation models. After fine-tuning a video diffusion model on a large corpus of videos paired with warped noise, we can control the motion of videos at inference time.

\subsection{\method noise warping}

\begin{algorithm}[t]
\caption{\method next-frame warping}
\label{alg:main}
\begin{algorithmic}[1]
\State \textbf{Input:} previous-frame noise $q \in \mathbb{R}^{D}$, previous-frame density $p \in \mathbb{R}^{D}$, forward flow $f: D \to \mathbb{N}^2$, backward flow $f': D \to \mathbb{N}^2$.
\State Let $G = (V, V', E)$ be a bipartite graph with $V = D$, $V' = D$ and edge set $E = \{\}$ to be constructed. 
\For{$v$ in $V$} \Comment{Contraction}
\State $E \gets E \cup (v, v+f(v))$ if $v+f(v')\in D$
\EndFor
\For{$v'$ in $V'$} \Comment{Expansion}
\If{$\deg_G(v') = 0$} \Comment{$\deg_G(v)$ denote the degree of $v$ in $G$}
\State $E \gets E \cup (v' + f'(v'), v')$ if $v' + f'(v')\in D$
\EndIf
\EndFor
\For{$v$ in $V$} \Comment{Conditional white noise sampling}
\State $d \gets \deg_G(v)$
\State Sample $Z \sim \mathcal{N}(0, I_{d})$, and set $S \gets \sum_{i=1}^d Z_i$
\State $X_i \gets \frac{q(v)}{d} + \frac{1}{\sqrt{d}}(Z_i - \frac{S}{d})$ for $i \in [d]$
\State $R(v) \gets \{X_i\}_{i\in[d]}$ 
\EndFor
\For{$(v')$ in $V'$} \Comment{Compute next-frame noise and density}
\State $q'(v') \gets 0$, $p'(v') \gets 0$, $s \gets 0$
\For{{$v$ in $V$ such that $(v,v') \in E$}}
\State $d \gets \deg_G(v)$, $\alpha \gets \frac{p(v)}{d}$
\State $q'(v') \gets q'(v') + \alpha R(v)\text{.pop}()$
\State $p'(v') \gets p'(v') + \alpha$
\State $s \gets s + \alpha^2 \frac{1}{d}$
\EndFor
\If{$s = 0$}
\State Sample $q'(v') \sim \mathcal{N}(0, 1)$
\Else
\State $q'(v') \gets \frac{q'(v')}{\sqrt{s}}$ \Comment{Renormalize to unit variance}
\EndIf
\EndFor
\State \textbf{return} next-frame noise and density $q', p'$.
\end{algorithmic}
\label{alg:algorithm}
\end{algorithm}

\subsubsection{Algorithm}

To facilitate the large-scale noise warping required by this new use case, we introduce a fast noise warping algorithm (\cref{alg:main}) that warps noise frame-by-frame, storing just the previous frame's noise (with dimensions $H \times W \times C$, where $H$ is height, $W$ is width, and $C$ is the number of channels) and a matrix of per-pixel flow density values (with dimensions $H \times W$). The density values indicate how much noise has been compressed into a given region. Unlike HIWYN~\cite{chang2024warped} which requires time-consuming polygon rasterization and upsampling of each pixel, our algorithm directly tracks the necessary \textit{expansion} and \textit{contraction} between frames according to the optical flow and uses only pixel-level operations that are easily parallelizable. We show that our algorithm retains the same Gaussianity guarantee as HIWYN~\cite{chang2024warped} (\cref{prop:gaussianity}).

\noindent \textbf{Next-frame noise warping}. Our noise warping algorithm calculates noise iteratively, where the noise for a given frame depends only on the state of the previous frame.

Let $H\times W$ be the dimensions of each video frame. Let $D = [H]\times [W]$ denote a 2D matrix with height $H$ and width $W$, where we use the notation $[n]:= {1,\ldots, n}$. Given the previous frame's noise\footnote{Since different channels are treated independently, we will assume a single channel in images.} $q \in \mathbb{R}^D$ and the flow density $p \in \mathbb{R}^D$ together with forward and backward flows\footnote{We allow flows to go out of bounds, i.e., $f$ and $f'$ can land in $\mathbb{N}^2 \setminus D$.} $f,f':D\to \mathbb{N}^2$, our algorithm computes the next-frame noise and density $q',p' \in \mathbb{R}^D$ such that $q'$ (resp. $p'$) is temporally correlated with $q$ (resp. $p$) via the flows.

At a high level, our algorithm (in~\cref{alg:algorithm}) combines two types of dynamics: \textit{expansion} and \textit{contraction}. In the case of \textit{expansion}, such as when a region of the video zooms in or an object moves towards the camera, one noise pixel is mapped to one or more noise pixels in the next frame (hence it ``expands''). In the case of \textit{contraction}, we adopt the Lagrangian fluid dynamics viewpoint of treating noise pixels as particles moving along the forward flow $f$. This often leaves gaps that need to be filled. Hence, for regions not reached when flowing along $f$, we use the backward flow $f'$ to pull back a noise pixel. That gap is filled with noise calculated with the \textit{expansion} case.

\begin{figure}
    \centering
    \includegraphics[width=\linewidth]
    {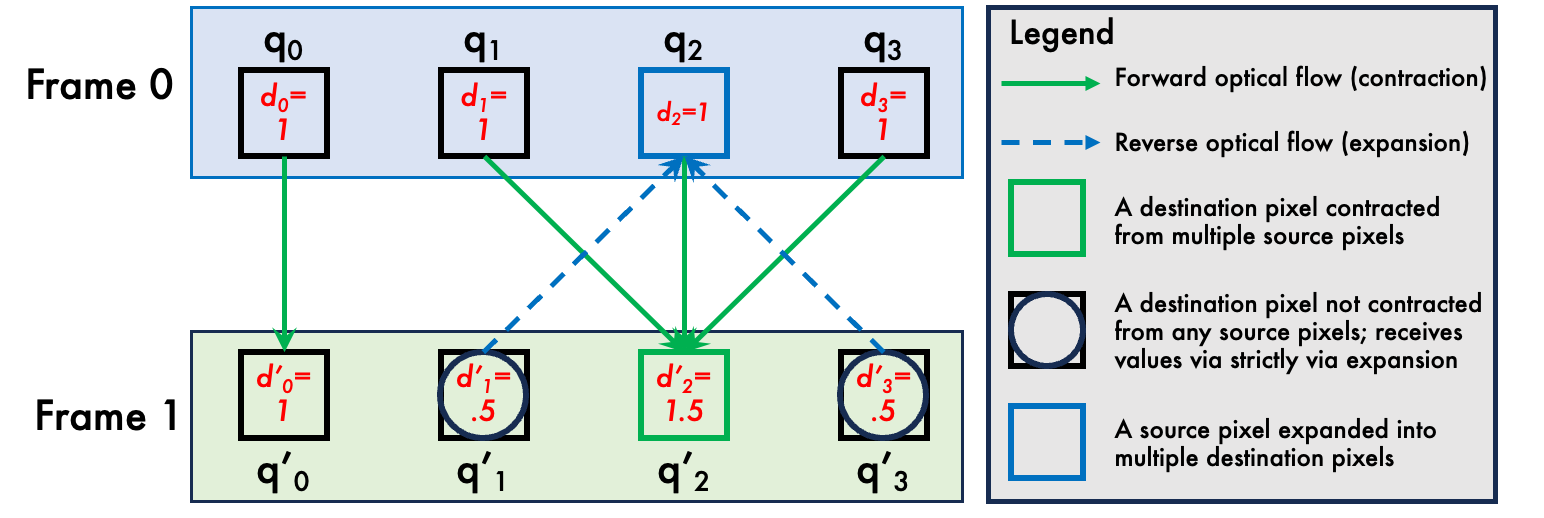}
    \caption{Diagram of our noise warping algorithm. A case example of our algorithm illustrates both the \textit{expansion} and \textit{contraction} cases, along with example density values. Each node represents some noise pixel `q'. Noise values $q_{0..3}$ are transferred from frame 0 to frame 1 using forward optical flow, and the remaining pixels in frame 1 that did not receive any values obtain their values from frame 0 using reverse optical flow (the \textit{expansion} case). In the \textit{contraction} cases such as $q'_2$, their densities become the sum of their sources. And in the \textit{expansion} case, where one source pixel spreads out into multiple target pixels, such as $q_2$ spreading out into $q'_1$ and $q'_3$, its density is dispersed.}
    \label{fig:algorithm_diagram}
\end{figure}

Additionally, to preserve the distribution correctly over long time periods, we use density values to keep track of how many noise pixels were aggregated into a given region, so that when mixed with other nearby particles in the \textit{contraction} case, these higher density particles have a larger weight. This is illustrated in \cref{fig:algorithm_diagram}.

We unify both \textit{expansion} and \textit{contraction} cases by building a bipartite graph $G$ where edges represent how noise and density should be transferred from the previous frame to the next. When aggregating the influence from graph edges to form the next-frame noise $q'$, we scale the noise in accordance with the flow density to ensure the preservation of the original frame's distribution, as detailed in ~\cref{alg:algorithm}. The \textit{expansion} and \textit{contraction} cases are calculated in tandem to prevent any cross-correlation, guaranteeing the output will be perfectly Gaussian.

\subsubsection{Theoretical analysis}

\begin{proposition}[Preservation of Gaussian white noise]
\label{prop:gaussianity}
If the pixels of the previous-frame noise $q$ in \cref{alg:main} are i.i.d. standard Gaussians, then the output next-frame noise $q'$ also has i.i.d. standard Gaussian pixels. Please check the appendix for a formal mathematical proof.
\end{proposition}

\begin{proposition}[Time Complexity]
\label{prop:time_complexity}
For a given frame, the time complexity of this algorithm is  $O(D)$, linear time with respect to the number of noise pixels processed. Proof: There are only two cases - \textit{contraction} and \textit{expansion}. Because each previous-frame pixel can only be contracted to one current-frame pixel, and during \textit{expansion} each current-frame pixel can only be mapped to one previous-frame pixel, the total number of edges $E$ will never exceed $2D$.
\end{proposition}

\subsection{Training-free image diffusion models with warped noise}

As shown by \citet{chang2024warped} and \citet{deng2024infinite}, noise warping can be combined with image diffusion models to yield temporally consistent video edits without training. To do this, we first take an input video and calculate its optical flows using RAFT~\cite{teed2020raft}. Then, with~\cref{alg:algorithm}, we use the flow fields to create sequences of Gaussian noise for each frame in the input video, ensuring that the noise moves along the flow fields. These noises are used during the per-frame diffusion processes in place of what would normally be temporally independently sampled Gaussian noise. This enables temporally consistent inference for video tasks, such as relighting~\cite{he2024diffrelight} and super-resolution~\cite{stabilityai2023deepfloyd}, using image-based diffusion models.

\begin{figure*}
    \centering
    \includegraphics[width=\linewidth]{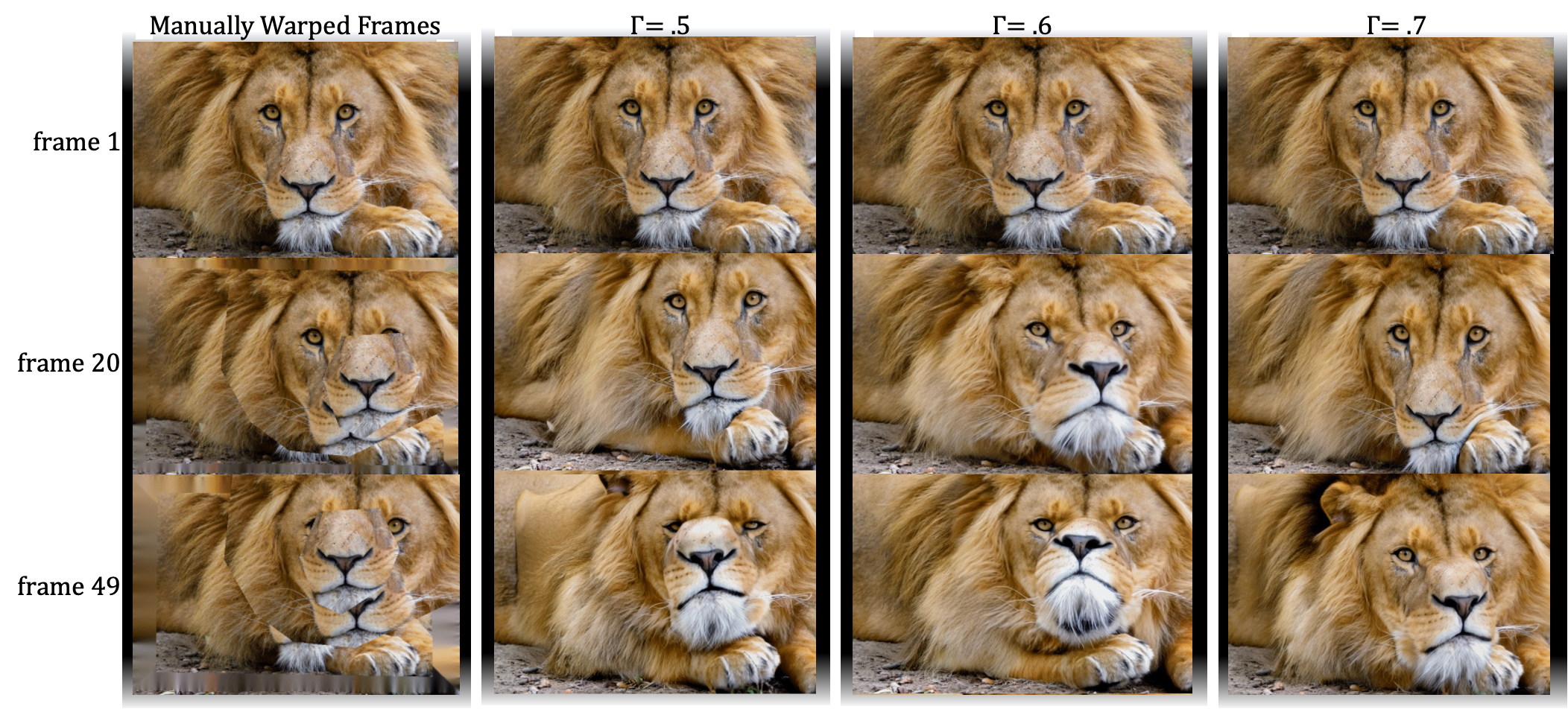}
    \caption{Showcasing the effect of noise degradation level $\deglevel$ on generated videos. A few frames from the driving video are shown in the leftmost column. Our model outputs are in the next 3 columns. As degradation decreases ($\deglevel$ from 0.7 to 0.5), the video more strictly adheres to the input flow. This allows us to control video movement with a user-specified level of precision.}
    \label{fig:degredation_lions}
\end{figure*}

\subsection{Fine-tuning video diffusion models with warped noise}

We use warped noise to condition a video diffusion model on optical flow. In particular, we fine-tune two variants of a latent video diffusion model CogVideoX~\cite{yang2024cogvideox}, both the text-to-video (T2V) and image-to-video (I2V) variants. We regard CogVideoX as a black box without changing its architecture.

We use the same training objective as in normal fine-tuning, i.e., the mean squared loss between denoised samples and samples with noise added. In fact, we use the exact same training pipeline as the original CogVideoX repository, with exactly one difference: during training, we use warped noise instead of regular Gaussian noise. For each training video, we calculate its optical flow for each frame, and create a warped noise tensor $\mathbf{Q} \in \mathbb{R}^{F \times C \times H \times W}$, where $F, C, H, W$ are the number of frames, the number of channels, the height and width of encoded video samples respectively by applying our algorithm iteratively.

We also introduce the concept of noise degradation, which lets us control the strength of our motion conditioning at inference time. After calculating the clean warped noise, we then degrade it by a random degradation level $\deglevel \in [0,1]$, by first sampling uncorrelated gaussian noise $\zeta \sim \mathcal{N}(0, 1)$ and modifying the warped noise $\Q \gets \frac{(1 - \deglevel) \Q + \zeta \deglevel}{\sqrt{(1 - \deglevel)^2 + \deglevel^2}}$. As degradation level $\deglevel \rightarrow 1$, $\Q$ approaches an uncorrelated Gaussian, and as $\deglevel \rightarrow 0$, $\Q$ approaches clean warped noise. At inference, the user can control how strictly the resulting video should adhere to the input flow. Please see \cref{fig:degredation_lions} for a qualitative depiction of the effect of $\deglevel$.

In practice, because the diffusion model works on latent embeddings, we calculate the optical flow and warped noise in image space and then downsample that noise into latent space, which in the case of CogVideoX means downscaling by a factor of $8\times8$ spatially and $4$ temporally. We use nearest-neighbor interpolation along the temporal axis and mean-pooling along the two spatial axes, which are then multiplied by $8$ to preserve unit variance.

\subsection{Video diffusion inference with warped noise}

At inference, we generate warped noise from an input video to guide the motion of the output video. Then, using a deterministic sampling process such as DDIM \cite{song2021denoising}, we use that warped noise to initialize the diffusion process of our fine-tuned video diffusion model. This method of control is much simpler than other motion control methods, as it does not require any changes to the diffusion pipeline or architecture - using exactly the same amount of memory and runtime as the base model.

In the case of local object motion control, we allow the user to specify object movements through a simple user interface as shown in~\cref{fig:comparisons_video_diffusion_object_motions}. It is used to generate synthetic optical flows, where multiple layers of polygons are overlaid on an image. Then, these polygons are translated, rotated and scaled with paths defined by the user. We warp the noise accordingly, and use that noise to initialize the diffusion process, along with a text prompt, and in the case of the image-to-video model, a given first frame image. By controlling the extent to which the output video follows these polygons, users can simulate camera movement by shifting the background, or even 3D motion effects by overlaying two polygons in parallax and moving them at different speeds. We find that this motion control representation is quite robust to user error, where even if the polygon only roughly matches the object or area of interest it will still produce high quality results. For synthetic object motion control, we typically use a degradation value $\deglevel$ between 0.5 and 0.7, depending on the level of motion precision the user desires, which is a higher level than we would normally use for motion transfer.

The case of motion transfer and camera motion control are very similar -- the only difference is the source of the flows used to generate the warped noise. In the case of motion transfer, we calculate the optical flow of a driving video, get warped noises that match the motion. Like in local object motion control, we use that warped noise to initialize a diffusion process. In the case of motion transfer, we typically use a lower degradation value $\deglevel$ between 0.2 and 0.5, as we usually want the output video's motion to match the driving video's motion as closely as possible.

\subsection{Implementation details}

We fine-tune the recent state-of-the-art open-source video diffusion model, CogVideoX-5B~\cite{yang2024cogvideox}, on both its T2V and I2V variants. We use a large general-purpose video dataset composed of 4M videos with resolution $\geq$720$\times$480 ranging from approximately 10 to 120 seconds in length, with paired texts captioned by CogVLM2~\cite{wang2024cogvlmvisualexpertpretrained}. We used 8 NVIDIA A100 80GB GPUs over the course of 40 GPU days, for 30,000 iterations using a rank-2048 LoRA~\cite{hu2021loralowrankadaptationlarge} with a learning rate of $10^{-5}$ and a batch size of 8.

Our method is data agnostic and model agnostic. It can be used to add motion control to arbitrary video diffusion models, while only processing the noise sampling during fine-tuning. For example, it also works with AnimateDiff \cite{guo2024animatediff} fine-tuned with the WebVid dataset~\cite{Bain21}, trained on 8$\times$40GB A100 GPUs over a period of 2 days with 12 frames and $256\times320$ resolution. See its qualitative results in \cref{fig:supp_animatediff_grid} in the supplementary material.
\begin{table*}[t]
    \centering
    \caption{Noise warping algorithm benchmarking in terms of Gaussianity, efficiency, and spatial quality and temporal consistency for two image diffusion based applications. $\Uparrow$/$\Downarrow$ indicates a higher/lower value is better.}
    \resizebox{0.9\textwidth}{!}{ 
    \begin{tabular}{l|cc|cccccccc}
        \toprule
        & \multicolumn{2}{c|}{\textbf{Noise w/o warping}} & \multicolumn{8}{c}{\textbf{ Noise warping method}} \\
        & Fixed & Random & Bilinear & Bicubic & Nearest & PYoCo & CaV & HIWYN & InfRes & Ours \\
        \midrule
        & \multicolumn{10}{c}{\textbf{Gaussianity}} \\
        Moran’s \textit{I} (index) $\Downarrow$ & -0.00027 & 0.00019 & 0.30 & 0.24 & 0.26 & 0.00023 & -0.00079 & 0.0011 &  0.00036 & 0.00014 \\
        Moran’s \textit{I} (p-value) $\Uparrow$ & 0.29 & 0.36 & 0.0 & 0.0 & 0.0 & 0.73 & 0.25 & 0.11 & 0.60 & 0.84 \\
        K-S Test (index) $\Downarrow$ & 0.089 & 0.075 & 0.34 & 0.37 & 0.17 & 0.13 & 0.073 & 0.062 & 0.055 & 0.060  \\
        K-S Test (p-value) $\Uparrow$ & 0.12 & 0.19 & 0.0005 & 0.0004 & 0.04 & 0.08 & 0.27 & 0.42 & 0.50 & 0.44 \\
        \midrule
        & \multicolumn{10}{c}{\textbf{Efficiency at 1024$\times$1024 resolution}} \\
        GPU time (ms) $\Downarrow$
        & $<$ 1 & $<$ 1 & 4.41 & 4.33 & 6.82 & 3.54 & 2.31 & 55.2 & 2.61 & 2.14 \\
        \midrule
                & \multicolumn{10}{c}{\textbf{Super-resolution - DeepFloyd IF}} \\
        LPIPS $\Downarrow$ & 0.29  & 0.29  & 0.60  & 0.62  & 0.55  & 0.28  & 0.28  & 0.29  & 0.28  & 0.29   \\
        SSIM $\Uparrow$ & 0.88  & 0.88  & 0.72  & 0.70  & 0.65  & 0.88  & 0.88  & 0.87  & 0.88  & 0.88  \\
        PSNR $\Uparrow$ & 29.36  & 29.41  & 28.68  & 28.55  & 28.59  & 29.40  & 29.39  & 29.31  & 29.38  & 29.39 \\
        Warping error $\Downarrow$ & 163.84  & 233.65  & 165.90  & 167.95  & 244.72  & 186.63  & 220.28  & 164.35  & 190.82  & 152.04 \\
        \midrule
        & \multicolumn{10}{c}{\textbf{Relighting - DifFRelight}} \\
        LPIPS $\Downarrow$ & 0.33 & 0.31 & 0.40 & 0.41 & 0.73 & 0.35 & 0.35 & 0.36 & 0.35 & 0.33 \\
        SSIM $\Uparrow$ & 0.69 & 0.77 & 0.73 & 0.70 & 0.38 & 0.58 & 0.67 & 0.64 & 0.60 & 0.70 \\
        PSNR $\Uparrow$ & 28.91 & 29.02 & 28.87 & 28.82 & 28.21 & 28.83 & 28.87 & 28.82 & 28.81 & 28.92 \\
        Warping error $\Downarrow$ & 86.65 & 128.11 & 47.53 & 43.57 & 164.42 & 95.24 & 106.77 & 87.72 & 87.97 & 85.82 \\
        \bottomrule
    \end{tabular}
    }
    \label{tab:comparisons_image_diffusion}
\end{table*}

\section{Experiments}
\label{sec:experiments}

\subsection{Gaussianity}

\textbf{Evaluation metrics}. To validate the preservation of spatial i.i.d. Gaussianity, we follow the evaluation protocol outlined by InfRes~\cite{deng2024infinite}. Specifically, we use Moran's \textit{I} to measure the spatial correlation of warped noise and the Kolmogorov-Smirnov (K-S) test to assess normality.

\noindent \textbf{Baselines}. Following HIWYN~\cite{chang2024warped}, we choose the per-frame fixed and independently-sampled noise as oracle baselines for perfect spatial Gaussianity but zero temporal correlation. We choose bilinear, bicubic, and nearest neighbor temporal interpolation as oracle baselines for sufficient temporal correlation but no spatial Gaussianity. We also compare with the recent noise warping algorithms including HIWYN~\cite{chang2024warped} and InfRes~\cite{deng2024infinite}. In line with these papers, we also include baselines \textit{Preserve Your Own Correlation} (PYoCo)~\cite{ge2023preserve} and \textit{Control-A-Video} (CaV)~\cite{chen2023control}, which have perfect Gaussianity but zero and insufficient temporal correlation, respectively.

\noindent  \textbf{Results}. According to \cref{tab:comparisons_image_diffusion} 1st section, we observe:

(1) For Moran's I, a value close to 0 indicates no spatial cross-correlation, which is desirable for i.i.d. noise. Our method achieves a Moran's I index of 0.00014 and a high p-value of 0.84, indicating strong evidence for no spatial autocorrelation. Similarly low Moran's I values and high p-values are observed for PYoCo, CaV, HIWYN and InfRes, because they also aim to generate spatially gaussian outputs.

(2) The K-S test compares the empirical distribution of the warped noise to a standard normal distribution. A small K-S statistic and a high p-value indicate the two distributions are similar. Our method obtains a K-S statistic of 0.060 and p-value of 0.44, suggesting the warped noise follows a normal distribution. Comparable results are seen for the other Gaussianity-preserving methods.

(3) In contrast, the bilinear, bicubic, and nearest neighbor warping methods fail to maintain Gaussianity, exhibiting Moran's I values an order of magnitude higher (0.24 to 0.30) with p-values of 0.0, and K-S statistics 3-6 times larger (0.17 to 0.37) with very low p-values ($<$0.05). These results provide strong evidence for the presence of spatial autocorrelation and deviation from normality in the warped noise from these interpolation-based methods.

\subsection{Efficiency}

Noise generation efficiency is measured by wall time profiling on an NVIDIA A100 40GB GPU, generating noise at a resolution of 1024$\times$1024 pixels. We compare with the same baselines as above. According to \cref{tab:comparisons_image_diffusion} 2nd section, our method runs faster than the concurrent InfRes and significantly outperforms the most recent published baseline HIWYN by $26\times$, due to our algorithm's linear time complexity. The efficiency is one order of magnitude faster than real time, validating our feasibility to apply noise warping on the fly during video diffusion model fine-tuning.

\subsection{Video editing via image diffusion}

To further validate the effectiveness of our noise warping algorithm, we repurpose off-the-shelf image-to-image diffusion models to perform video-to-video editing tasks in a frame-by-frame manner, without training. Noise is warped using our algorithm and the above baselines based on the RAFT optical flow~\cite{teed2020raft} from input video and fed to two image pre-trained diffusion models: DeepFloyd IF~\cite{stabilityai2023deepfloyd} for super-resolution and DifFRelight~\cite{he2024diffrelight} for portrait relighting. By measuring the quality and temporal consistency of the output video, we can effectively evaluate the spatial Gaussianity and temporal consistency of different noise warping algorithms.

\noindent \textbf{Evaluation metrics}. We use LPIPS~\cite{zhang2018unreasonable}, SSIM~\cite{hore2010image}, and PSNR~\cite{hore2010image} to measure the quality of the output frames w.r.t. ground truth frames. We use warping error~\cite{lai2018learning} to measure temporal consistency (mean square error) between two adjacent generated frames after flow warping.

\subsubsection{DeepFloyd IF video super-resolution}

We evaluate noise warping on DeepFloyd~IF~\cite{stabilityai2023deepfloyd} super-resolution using 43 videos from the DAVIS dataset~\cite{pont20172017}. The videos were downsampled to the 64$\times$64 and super-resolved to 256$\times$256.

\noindent \textbf{Results}. According to \cref{tab:comparisons_image_diffusion} 3rd section, our algorithm outperforms all the baselines in terms of temporal consistency (warping error). Our supplementary video also shows that our algorithm is more stable for the foreground, background, and edges, in contrast to InfRes which is often unstable in the background and HIWNY which is much less stable around moving edges. Our algorithm is comparable to other methods in PSNR, SSIM, and LPIPS image quality metrics, apart from the bilinear, bicubic, and nearest methods which result in low quality generation due to spatial non-Gaussianity. See \cref{fig:supp_davis_deepfloyd} in the supplementary material for more details.

\subsubsection{DifFRelight video relighting}

We evaluate noise warping on DifFRelight~\cite{he2024diffrelight} portrait video relighting using their own dataset, which includes 4 subjects in 4 scenarios: a 180-degree view animation, a 720-degree view animation, a zigzag camera movement sequence, and an interpolating camera path through several fixed stage capture positions, all with fixed lighting conditions. During inference, we center crop a 1024$\times$1024 region out of a 1080$\times$1920 Gaussian splat rendering and infer with various noises using conditioned lighting.

\noindent \textbf{Results}. According to \cref{tab:comparisons_image_diffusion} 4th section, throughout all baseline comparisons, our algorithm shows consistently advantageous scores in both image and temporal metrics, validating its fundamental benefits to the image diffusion model. Although our visual results at first glance are comparable to HIWYN and InfRes in the supplementary \cref{fig:supp_diffrelight_noisewarp} and our \href{https://eyeline-research.github.io/Go-with-the-Flow/}{webpage}, its visual improvements can be seen in the beard regions and skin reflections. We also notice quite low warping error values on the bilinear and bicubic noise inferences, likely coming from the long blurry streaks generated along the flow, while at the same time image quality deteriorates significantly.

\begin{figure}
    \centering
    \includegraphics[width=\linewidth]{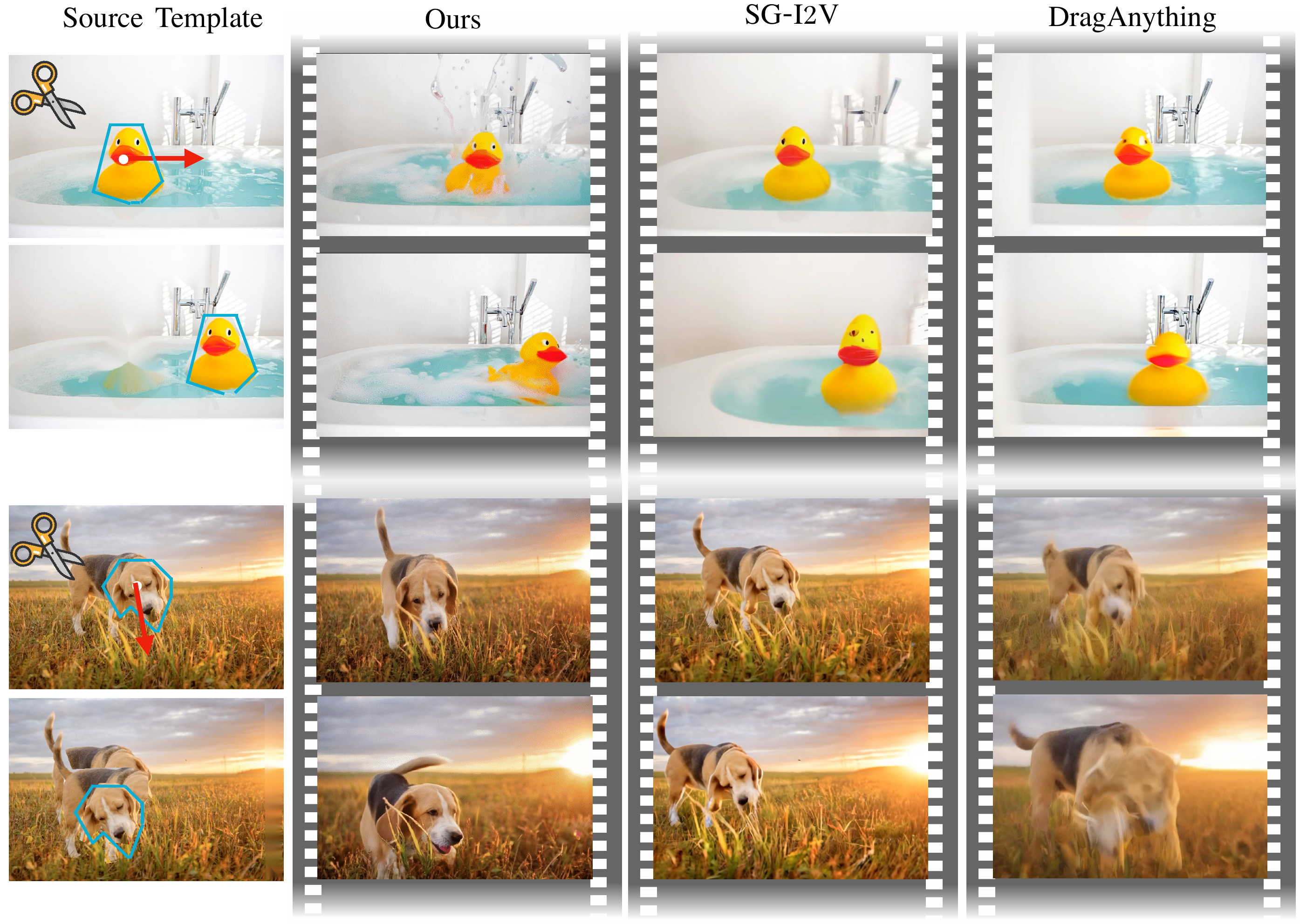}
    \caption{Qualitative comparisons of local object motion control. Zoom in for details. The user selects any number of polygons, then scales, rotates, or translates them along arbitrary paths, which are then used to create the warped noise flow.}
    \label{fig:comparisons_video_diffusion_object_motions}
\end{figure}

\begin{figure}
    \centering
    \includegraphics[width=\linewidth]{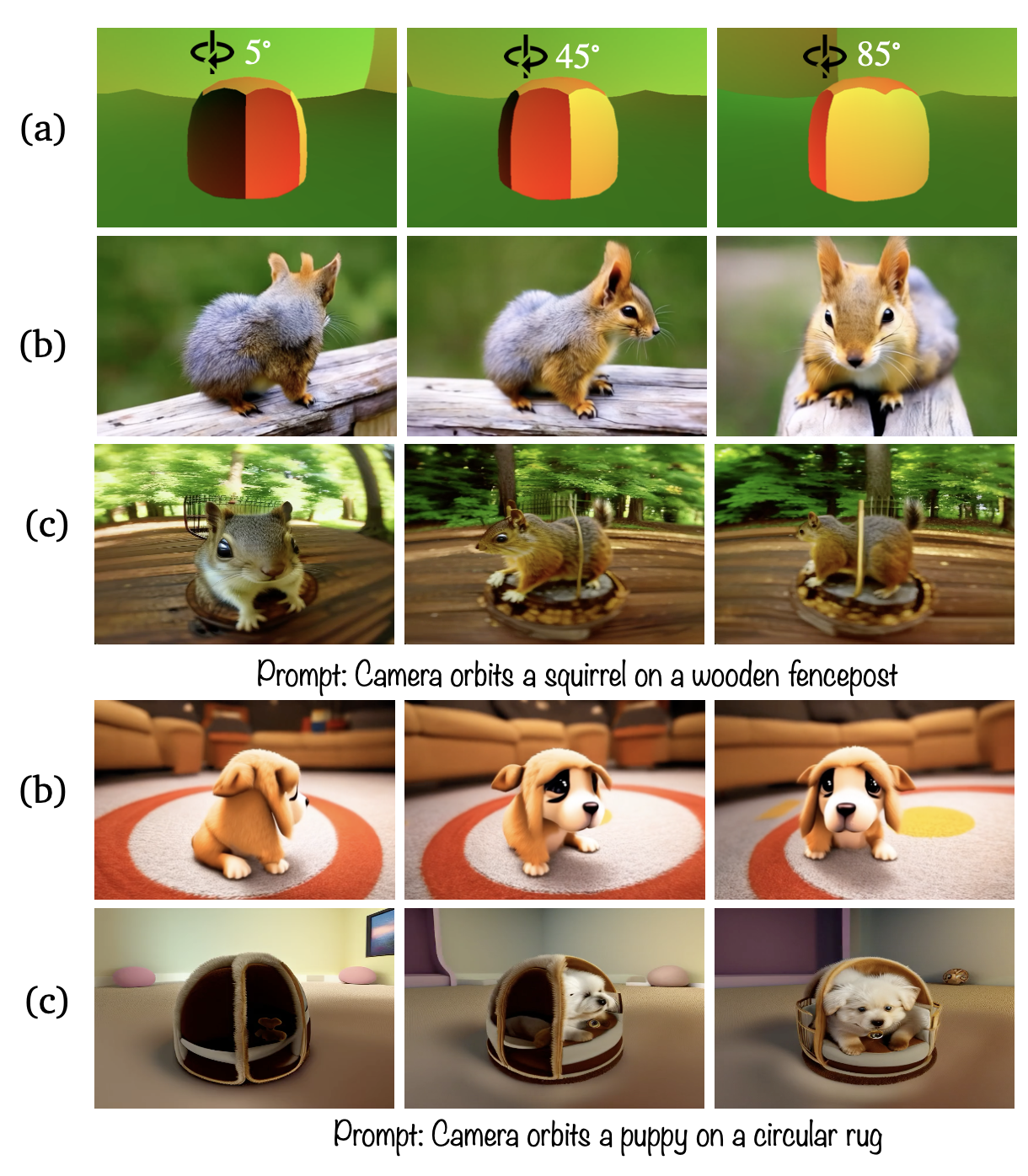}
    \caption{Qualitative comparisons of camera movement video generation of our method (b) and MotionClone (c) using a turning source video (a).}
    \label{fig:comparisons_video_diffusion_turning_object}
\end{figure}

\subsection{Video diffusion with motion control}

\subsubsection{Local object motion control}

We introduce a novel method for controlling object motion, by leveraging the flows of input templates. These templates include user-defined local region masks and cut-and-drag trajectories that allow users to specify the motion of one or more objects built with a simple, intuitive UI (\cref{fig:comparisons_video_diffusion_object_motions}), and synthetic flows of a camera rotating around 3D objects (\cref{fig:comparisons_video_diffusion_turning_object}). 

During inference, we use the precise flow computed from the input template frames to guide noise warping for video generation. This enables our I2V model to apply accurate, localized movements and adjustments to the input image while preserving object structure and faithfully following the intended motion trajectory.

\RYAN{
We also provide quantitative benchmarks. Following \citep{namekata2024sg}, we use the VIPSeg \cite{miao2021vspw} to benchmark our method on local object motion control, as well as the 40 videos from our user study.
}

\noindent \textbf{Baselines}. We evaluate our video generation model against five state-of-the-art baselines, SG-I2V~\cite{namekata2024sg}, MotionClone~\cite{ling2024motionclone}, 
DragAnything~\cite{wu2024draganything}, to benchmark its ability to accurately control object and camera movements derived from a given input template. One of the most recent works, SG-I2V, is an I2V model for object motion transfer guided by bounding box trajectories. We adapt our user-defined polygons to bounding boxes as its input.

\noindent \textbf{Results}. From ~\cref{fig:comparisons_video_diffusion_object_motions}, \cref{fig:comparisons_video_diffusion_turning_object}, \cref{tab:comparisons_video_diffusion_motion_transfer} and our \href{https://eyeline-research.github.io/Go-with-the-Flow/}{webpage}, we observe:

(1) Existing methods struggle to handle complex, localized object motions. Specifically, when specifying local adjustments, such as rotating a dog's head while keeping the rest of the body static, these methods often fail, applying unnatural translational or global transformations to the entire object.

(2) We find that SG-I2V frequently misinterprets object-specific movements as global camera shifts, resulting in scene-wide translations rather than accurate object manipulations.

(3) DragAnything, which employs single-line trajectory control, lacks temporal and 3D consistency, leading to significant distortions and reduced fidelity in complex motion scenarios.

(4) MotionClone also fails to capture subtle object dynamics, as it relies on sparse temporal attention for motion guidance and is likely limited by the low spatial resolution of its diffusion features.

(5) Qualitatively, our model outperforms these baselines by maintaining high object fidelity and 3D consistency, even in scenarios with intricate or overlapping motions. Notably, our approach preserves object integrity and introduces plausible physical interactions, such as generating realistic splashes when moving a duck within a tub. Extensive user studies and quantitative evaluations validate our superior performance in motion consistency, visual fidelity, and overall realism.

(6) Our quantitative evaluation matches our qualitative observations. On both VIPSeg and the 40 videos from our user-study, our method outperforms all the training-based and training-free baselines.

\noindent \textbf{User study}. We conducted a comprehensive user study with 40 participants, asking them to evaluate and rate different methods based on their effectiveness in object motion control and maintaining 3D and temporal consistency. Our method stands out significantly, achieving a win percentage of \textbf{82\%} for \cutndrag local object motion control like~\cref{fig:comparisons_video_diffusion_object_motions} and \textbf{90\%} for the turnable camera movement control like~\cref{fig:comparisons_video_diffusion_turning_object}. The three baselines have substantially lower performance levels. More user study details are included in the supplementary material \cref{sec:supp_user_study} and \cref{fig:supp_user_study_screenshots_statistics}.

\begin{figure}
    \centering
    \includegraphics[width=\linewidth]{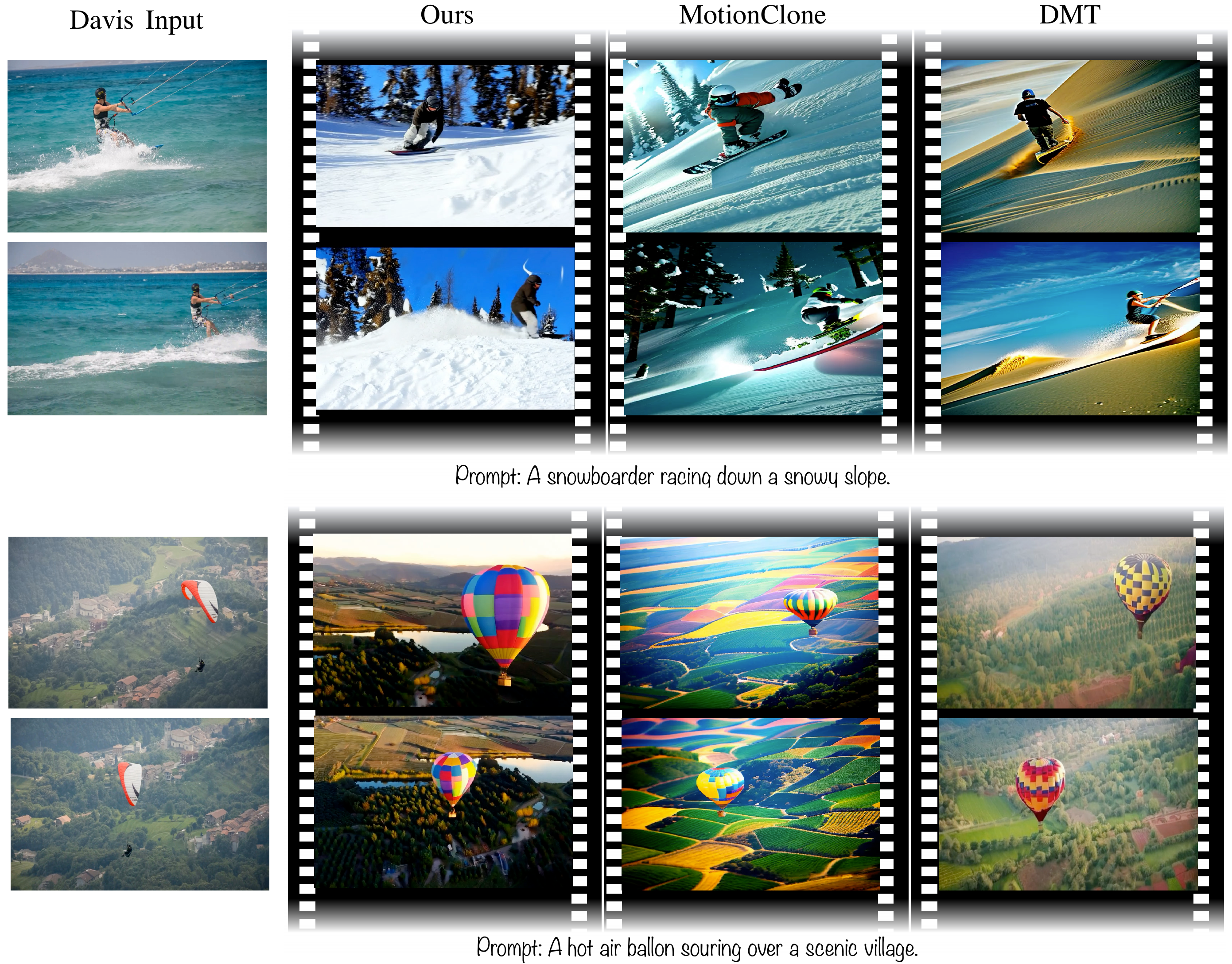}
    \caption{Qualitative comparisons of motion transfer T2V on the DAVIS dataset. Zoom-in needed.}
    \label{fig:comparisons_davis_i2v}
\end{figure}

\begin{figure}
    \centering
    \includegraphics[width=\linewidth]{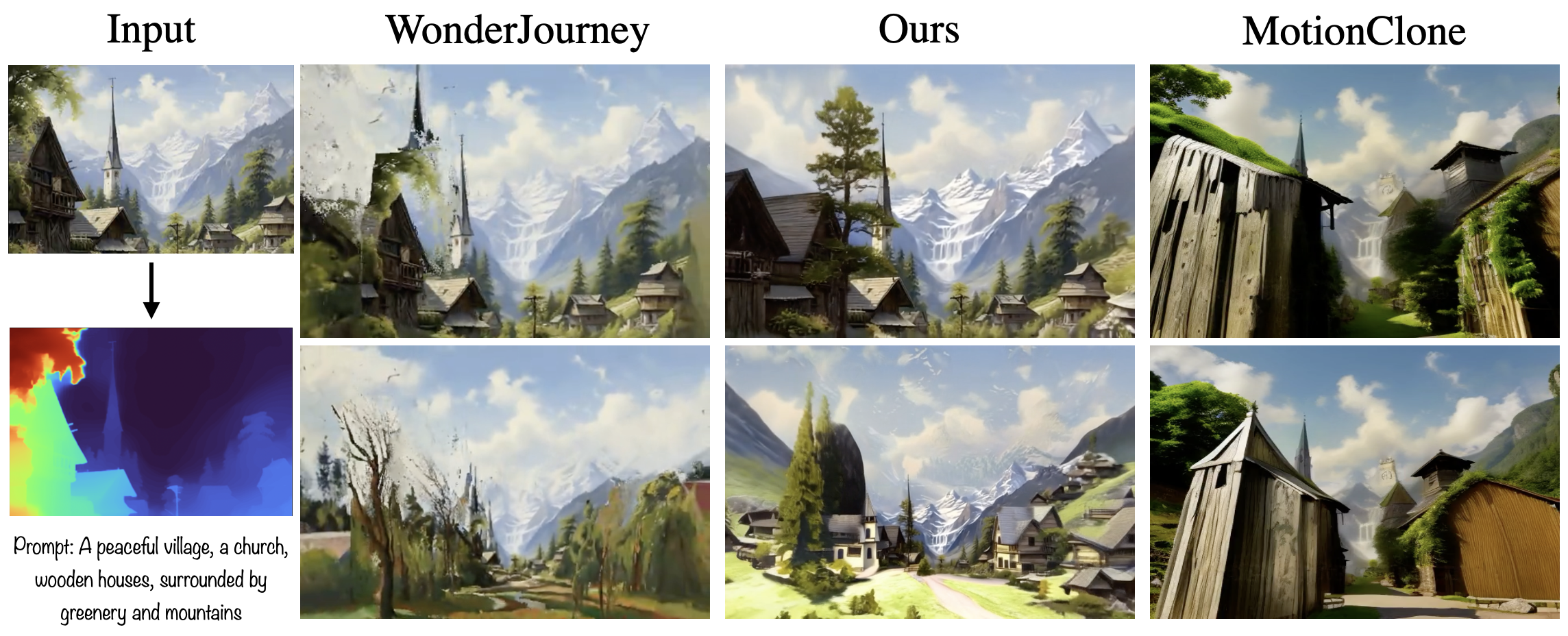}
    \caption{We apply our method to a sequence of frames warped using monocular depth estimation, enabling consistent 3D scene generation from a single image. In this example, we use results from WonderJourney. Zoom-in needed.
    }
    \label{fig:comparisons_video_diffusion_WonderJourney}
\end{figure}

\begin{figure}
    \centering
    \includegraphics[width=1\linewidth]{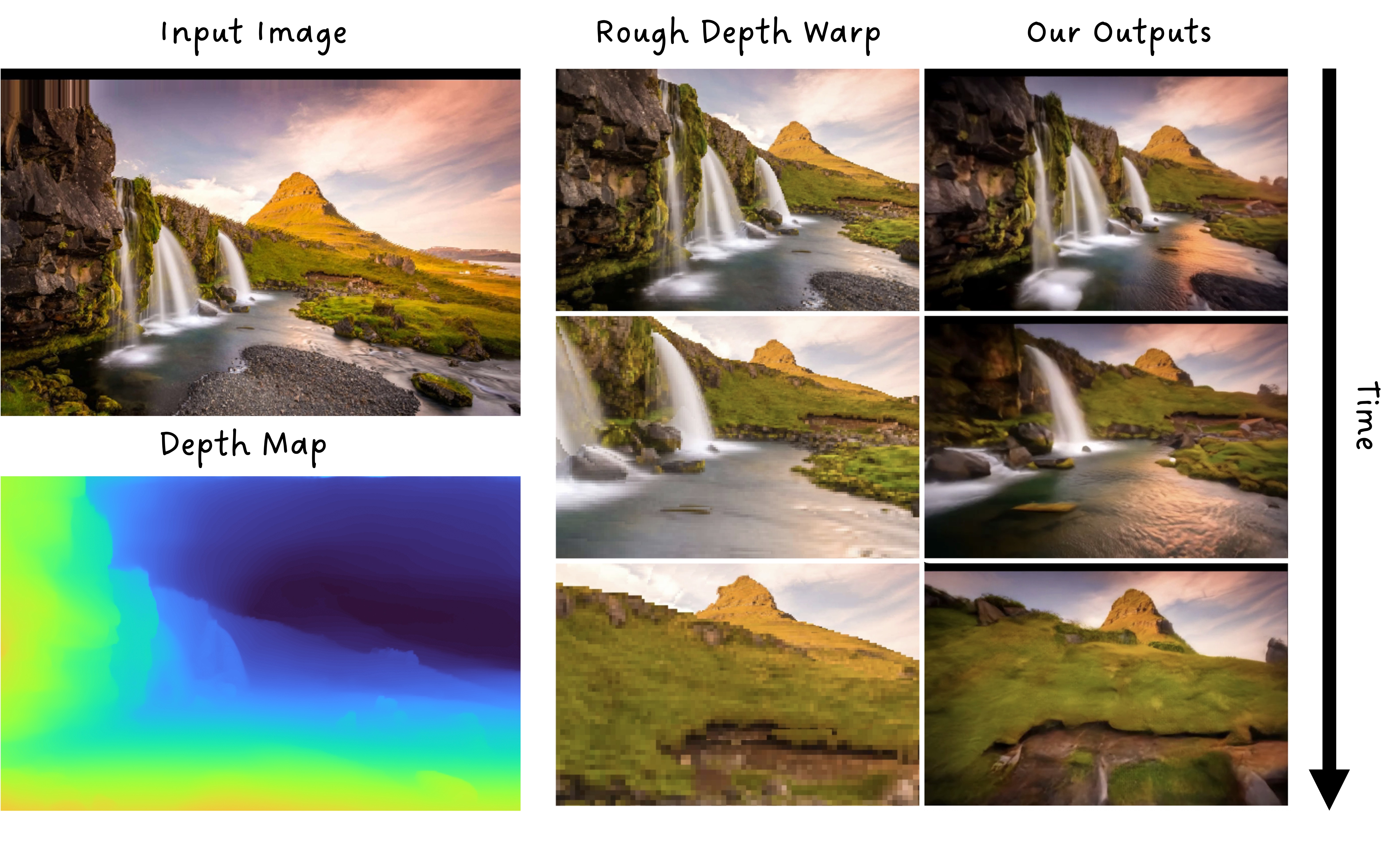}
    \caption{We explore a 3D scene, flying into a given image. Similar to \cref{fig:comparisons_video_diffusion_WonderJourney}, we take an image as an input and use a monocular depth estimator DepthPro \cite{bochkovskii2024depthprosharpmonocular} to get a depth map. Then, we use that depth to generate a crudely warped video (note the pixelation on the rough depth warp column when zoomed in) - and from the movement in that video get warped noise. From there, we run our motion-conditioned I2V model.}
    \label{fig:depth_warp}
\end{figure}

\begin{figure}
    \centering
    \includegraphics[width=1\linewidth]{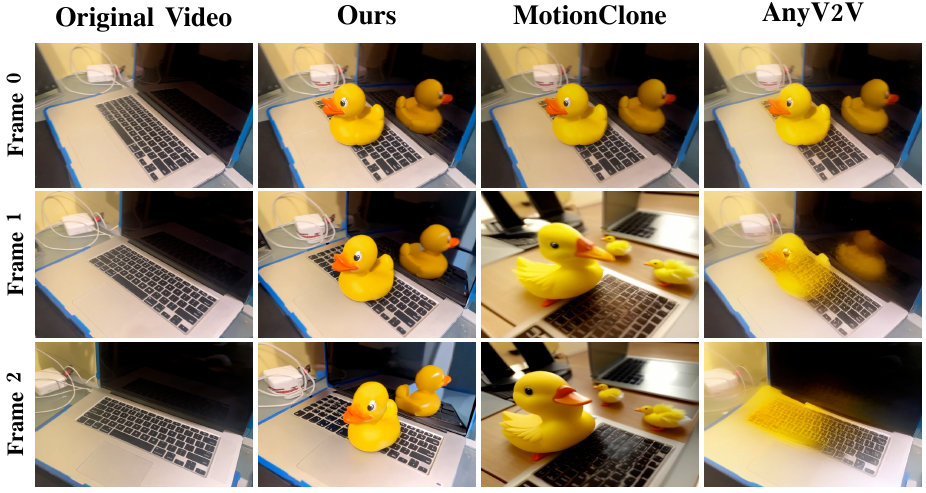}
    \caption{Comparison of initial frame video editing results across different methods. All methods start with the same edited initial frame derived from the original video.}
    \label{fig:init_frame_edit}
\end{figure}

\subsubsection{Motion transfer and camera movement control}

Our method also supports motion transfer and camera movement control, working with both T2V and I2V video diffusion models. By using reference videos and applying noise warping based on their optical flows, it can effectively capture and transfer complex motions.

\noindent \textbf{Datasets}. We choose the DAVIS video dataset~\cite{pont20172017} containing 43 videos of general object motion with ground truth object segmentation annotations, a random subset of 100 videos from the DL3DV dataset~\cite{ling2024dl3dv}, and 19 videos generated with WonderJourney~\cite{yu2024wonderjourney} that predominantly feature camera movements (\cref{fig:comparisons_video_diffusion_WonderJourney}), which itself uses depth-warping.

\noindent \textbf{Evaluation metrics}. For pixel quality, we calculate Fréchet Inception Distance (FID) between a set of real and generated frames. For motion controllability, we calculate (1) the mean Interaction over Union (mIoU) of CoTracker's tracking bounding boxes~\cite{karaev2023cotracker} between ground-truth and generated videos, and (2) the pixel MSE between ground-truth and generated videos, considering an I2V diffusion model is conditioned on ground-truth prompts, ground-truth initial frames, and ground-truth motion trajectories/flows. For text controllability, we calculate the cosine similarity between the prompt's CLIP~\cite{radford2021learning} text embedding and the generated frames' CLIP image embeddings, and average over frames of a generated video. For temporal consistency, we calculate (1) the cosine similarity of the CLIP image embeddings between two consecutive generated frames and average over all pairs in a generated video, and (2) the Fréchet Video Distance (FVD)~\cite{unterthiner2018towards} between a set of real and generate videos. In addition, we also benchmark on four metrics of VBench~\cite{huang2024vbench}, specifically for the temporal consistency/smoothness dimension.

\begin{table*}
    \centering
    \caption{Quantitative comparisons of motion transfer. $\Uparrow$/$\Downarrow$ indicates a higher/lower value is better. \textbf{Bold} indicates the best results. \colorbox{gray!20}{Gray background rows} indicate our final model. Dashed lines separate ablation study from baseline benchmarking.}
    \definecolor{Gray}{gray}{0.8}
    \resizebox{\linewidth}{!}{
        \begin{tabular}{l|c|ccccccccccc}
        \toprule
        & & & CoTracker & {Optical} & Pixel & CLIP & CLIP & FVD & \multicolumn{4}{c}{VBench $\Uparrow$} \\
        & {Training} & FID & mIoU & {flow} & MSE & text & image & $\times 10^3$ & Subject & Background & Motion & Temperal \\
        & {free?} & $\Downarrow$ & $\Uparrow$ & {err.} $\Downarrow$ & $\Downarrow$ & $\Uparrow$ & $\Uparrow$ & $\Downarrow$ & consistency & consistency & smoothness & flickering\\
        \midrule
        & & \multicolumn{11}{c}{{\textbf{Local object motion control on VIPSeg}}}\\
        MotionClone & \checkmark & 85.2 & 0.71 & 0.48 & 0.086 & 0.31 & 0.95 & 1.26 & 0.88 & 0.85 & 0.94 & 0.90\\
        SG-I2V & \checkmark & 61.4 & 0.63 & 0.84 & 0.065 & 0.31& 0.97 & 1.06 & \textbf{0.93} & \textbf{0.95} & 0.96 & 0.94\\
        \rowcolor{Gray} Ours & $\times$ & \textbf{41.1} & \textbf{0.75} & \textbf{0.36} & \textbf{0.039} & \textbf{0.32} & \textbf{0.98} & \textbf{0.47} & 0.91 & 0.92 & \textbf{0.97} & \textbf{0.95} \\ 
        \midrule
        & & \multicolumn{11}{c}{{\textbf{Local object motion control on our 40 samples in the user study}}}\\
        MotionClone & \checkmark & 96.6 & - &  0.80 & 0.048 & \textbf{0.33} & \textbf{0.98} & 1.38 & 0.86 & 0.93 & 0.97 & 0.95 \\
        SG-I2V & \checkmark & 79.9 & - &  0.64 & 0.042 & 0.32 & \textbf{0.98} & 1.27 & 0.95 & \textbf{0.95} & \textbf{0.98} & 0.94 \\
        DragAnything & $\times$ & 82.8 & - & 0.62 & 0.047 & 0.31 & 0.97 & 1.30 & 0.93 & \textbf{0.95} & \textbf{0.98} & 0.95 \\
        \rowcolor{Gray} Ours & $\times$ & \textbf{74.3} & - & \textbf{0.56} & \textbf{0.028} & 0.32 & \textbf{0.98} & \textbf{0.94} & \textbf{0.96} & \textbf{0.95} & \textbf{0.98} & \textbf{0.96} \\
        \midrule
        & & \multicolumn{11}{c}{\textbf{Motion transfer T2V on DAVIS}}\\
        DMT & \checkmark & - & \textbf{0.85} & \textbf{0.28} & - & 0.31 & 0.95 & - & 0.86 & 0.92 & 0.94 & \textbf{0.91} \\
        MotionClone & \checkmark & - & 0.75 & 0.38 & - & 0.32 & 0.93 & - & 0.78 & 0.89 & 0.86 & 0.81\\
        {MotionCtrl} & $\times$ & & 0.47 & 0.85 & -& 0.32 & 0.97 & - & 0.97 & 0.93 & \textbf{0.98} & 0.92  \\
        \rowcolor{Gray} Ours & $\times$ & - & 0.70 & 0.41 & - & \textbf{0.33} & \textbf{0.98} & - & 0.88 & \textbf{0.93} & 0.97 & 0.89\\
        \hdashline
        {Ours-CogVideoX-2B} & $\times$ & - & 0.64 & 0.48 & - & 0.32 & 0.95 & - & 0.89 & 0.91 & 0.97 & 0.90\\
        \midrule
        & & \multicolumn{11}{c}{\textbf{Motion transfer I2V on DAVIS}}\\
        MotionClone & \checkmark & 99.4 & 0.72 & 0.42 & 0.068 & 0.31 & 0.94 & 1.84 & 0.75 & 0.85 & 0.92 & 0.87 \\
        {ImageConductor} & $\times$ & 104.6 & 0.66 & 0.64 & 0.072 & 0.31 & 0.93 &  1.58 & 0.77 & 0.88 & 0.93 & 0.90 \\
        Original CogVideoX-5B & \checkmark & \textbf{76.62} & 0.52 & 0.67 & 0.088 & 0.31 & 0.96 & 1.36 & 0.85 & 0.91 & 0.96 & 0.92 \\
        \rowcolor{Gray} Ours ($\gamma=0.5$) & $\times$ & 78.6 & \textbf{0.74} & \textbf{0.36} & \textbf{0.053} & 0.31 & \textbf{0.97} & \textbf{1.21} & \textbf{0.88} & \textbf{0.92} & \textbf{0.98} & \textbf{0.93}\\
        \hdashline
        {Our ($\gamma=0.9$)} & $\times$ & 92.5 & 0.50 & 0.65 & 0.072 & 0.31 & 0.95 &  1.59 & 0.80 & 0.89 & 0.94 & 0.91\\
        {Our ($\gamma=0.8$)} & $\times$ & 80.6 & 0.68 & 0.47 & 0.067 & 0.31 & 0.96 & 1.50 & 0.85 & 0.91 & 0.96 & 0.92 \\
        {Our ($\gamma=0.4$)} & $\times$ & 77.7 & 0.74 & \textbf{0.36} & 0.056 & 0.31 & 0.97 & 1.27 & 0.87 & 0.91 & 0.97 & \textbf{0.93}\\
        {Our ($\gamma=0.2$)} & $\times$ & 77.1 & 0.74 & 0.37 & 0.058 & \textbf{0.32} & 0.97 & 1.29 & 0.86 & 0.91 & 0.97 & \textbf{0.93}\\
        \hdashline
        {Our (33\% data)} & $\times$ & 100.1 & 0.73 & 0.40 & 0.066 & 0.31 & 0.97 & 1.46 & 0.85 & 0.90 & 0.97 & 0.92\\
        {Our (12.5\% data)} & $\times$ & 105.2 & 0.71 & 0.39& 0.072 & 0.31 & 0.96 & 1.93& 0.84 & 0.89 & 0.97 & 0.91\\
        \midrule
        & & \multicolumn{11}{c}{\textbf{Camera movement transfer I2V on DL3DV}}\\
        MotionClone & \checkmark & 82.7 & 0.71 & 0.44 & 0.104 & \textbf{0.33} & 0.94 & 1.11 & 0.74 & 0.85 & 0.91 & 0.86 \\
        {ImageConductor} & $\times$ & 89.2 & 0.61 & 0.78 & 0.068 & 0.31 & 0.95 & 0.91 & 0.85 & 0.90 & 0.95 & \textbf{0.93}\\
        \rowcolor{Gray} Ours & $\times$ & \textbf{48.4} & \textbf{0.83} & \textbf{0.20} & \textbf{0.046} & 0.32 & \textbf{0.97} & \textbf{0.34} & \textbf{0.88} & \textbf{0.92} & \textbf{0.97} & \textbf{0.93}\\
        \midrule
        & & \multicolumn{11}{c}{\textbf{Camera movement transfer I2V on WonderJourney}}\\
        MotionClone & \checkmark & 177.9 & 0.81 & 0.17 & 0.103 & \textbf{0.32} & 0.96 & 1.93 & 0.75 & 0.87 & 0.93 & 0.87 \\
        {ImageConductor} & $\times$ & 166.1 & 0.79 & 0.39 & 0.085 & \textbf{0.32} & 0.94 & 1.63 & 0.79 & 0.88 & 0.93 & 0.90 \\
        \rowcolor{Gray} Ours & $\times$ & \textbf{128.3} & \textbf{0.85} & \textbf{0.15} & \textbf{0.072} & 0.31 & \textbf{0.98} & \textbf{1.55} & \textbf{0.82} & \textbf{0.91} & \textbf{0.98} & \textbf{0.94}\\
        \bottomrule
    \end{tabular}
    }
    \label{tab:comparisons_video_diffusion_motion_transfer}
\end{table*}

\noindent \textbf{Baselines}. For the motion transfer T2V scenario, we compare with the recent state-of-the-art methods Diffusion Motion Transfer (DMT)~\cite{yatim2024space}, MotionClone~\cite{ling2024motionclone}, and \RYAN{MotionCtrl~\cite{wang2024motionctrl}}. For the motion transfer I2V, we compare MotionClone and \RYAN{ImageConductor~\cite{li2024imageconductorprecisioncontrol}}) as DMT does not take an image as input.

In addition, we demonstrate video first-frame editing, a challenge where a user starts with an original video and an edited version of its initial frame. The goal is to seamlessly propagate the edits made to the first frame throughout the entire video while preserving the original motion. We qualitatively compare with MotionClone~\cite{ling2024motionclone} and the state-of-the-art video editing method AnyV2V~\cite{ku2024anyv2v} on real videos with photoshopped first frames.

We also source a few images for image-based depth warping, where we take an image, use a monocular depth estimator, DepthPro~\cite{bochkovskii2024depthprosharpmonocular}, to get a depth map, and crudely warp it to simulate a desired camera trajectory.

\noindent \textbf{Results}.
We present both qualitative and quantitative comparisons with baselines in~\cref{tab:comparisons_video_diffusion_motion_transfer}, \cref{fig:comparisons_davis_i2v}, \cref{fig:comparisons_video_diffusion_WonderJourney}, \cref{fig:depth_warp}, \cref{fig:init_frame_edit}, and our \href{https://eyeline-research.github.io/Go-with-the-Flow/}{webpage}. We observe:

(1) Our superior object motion transfer: On the DAVIS dataset, which includes object motion along with some degree of camera movement, our method demonstrates improved motion fidelity and overall video quality, as measured by Vbench. In particular, in the I2V setting, where both the initial frame and the source video are provided, our method achieves significantly better scores in FID, FVD, and motion metrics, indicating much closer reconstructions of the ground truth videos.

(2) Our superior camera movement control: On the DL3DV and WonderJourney datasets, which involve substantial camera movement, our method notably outperforms MotionClone in both motion fidelity and general video quality. This highlights our method's ability to effectively replicate intricate camera movements while maintaining visual coherence. For our depth-warping example~\cref{fig:depth_warp}, our results are far better than simply warping an image from its depth map, resulting in a smooth, realistic camera trajectory. See our \href{https://eyeline-research.github.io/Go-with-the-Flow/}{webpage} for videos.

(3) Our superior video first-frame editing: In~\cref{fig:init_frame_edit}, our method seamlessly integrates the added object into the scene while accurately preserving the camera \ning{movement} from the original video. In contrast, both baselines exhibit significant identity loss: MotionClone generates additional, unintended objects, and in AnyV2V, the foreground object gradually disappears. This demonstrates the superiority of our method in maintaining the original video's motion while faithfully preserving the identity of the object added to the first frame.

\subsubsection{Ablation studies}

In~\cref{tab:comparisons_video_diffusion_motion_transfer} for the DAVIS I2V task, we compare our method with a variant that excludes motion conditioning using warped noise (``Original CogVideoX-5B''), relying solely on textual prompts describing the objects. We observe: (1) Better video reconstruction: Our method, which incorporates motion conditioning, achieves superior FID, FVD, and CoTracker mIoU scores, indicating more accurate reconstruction of the source video. This is because textual prompts and the initial frame alone are insufficient to capture a video's future dynamics, whereas incorporating real video-derived motion guidance enables the generation of more realistic sequences. (2) Improved video quality: By utilizing warped noise for motion conditioning, our approach not only maintains but also enhances overall video quality, as measured by Vbench, demonstrating that integrating realistic motion cues improves the plausibility of the generated videos without compromising quality.



\RYAN{
Further exploring the effects of degradation, we find that as the degradation values $\gamma$ increase, the motion control becomes tighter - resulting in higher optical flow and CoTracker-mIoU scores, along with a closer per-frame similarity to target videos on the I2V DAVIS task. We find that, in general, $\gamma\approx 0.5$ is a good value for most tasks.}


\RYAN{We also perform ablations on dataset size, comparing models trained on different fractions of our dataset: training with a fraction of dataset yields worse performance than our full model.}

\RYAN{In addition, we perform an ablation where we use the weaker base model CogVideoX-2B, and find its performance is weaker than our main T2V model, based on CogVideoX-5B.}
\section{Conclusion}
\label{sec:conclusion}

In this work, we introduce a novel and faster-than-real-time noise warping algorithm that seamlessly incorporates motion control into video diffusion noise sampling, bridging the gap between chaos and order in generative modeling. By leveraging this noise warping technique to preprocess video data for video diffusion fine-tuning, we provide a unified paradigm for a wide range of user-friendly, motion-controllable video generation applications. Extensive experiments and user studies demonstrate the superiority of our method in terms of visual quality, motion controllability, and temporal consistency, making it a robust and versatile solution for motion control in video diffusion models.

\section*{Acknowledgments}
\label{sec:acknowledgment}

We would like to express our gratitude to Stephan Trojansky and Jeffrey Shapiro for their initial and ongoing executive support; Sebastian Sylwan, Daniel Heckenberg, Jitendra Agarwal, Matheus Leão, and Sungmin Lee for their IT support; Xueming Yu and David George for their hardware support; Jennifer Lao and Lianette Alnaber for their operational support; and Winnie Lin, Ahmet Tasel, Yiqun Mei, Lukas Lepicovsky, Rahul Garg, Ashish Rastogi, Ritwik Kumar, Cornelia Carapcea, and Girish Balakrishnan for their insightful technical discussions.

\section*{Social impact statement}
\label{sec:social_impact_statement}

Our work contributes to the growing field of video generative models by advancing motion-controllable video generation, which has the potential to revolutionize creative industries such as filmmaking and animation. By introducing a computationally efficient and accessible framework, our method democratizes high-quality video generation, enabling creators, developers, and artists to produce dynamic content with minimal resources or specialized training.

However, we acknowledge the potential misuse of such technology, including the creation of deepfakes or misleading media. To mitigate these risks, we advocate for responsible use, proper content labeling, and the integration of detection mechanisms to ensure ethical deployment. Our approach also emphasizes compatibility with diverse models, encouraging transparency and collaboration within the research community to address societal concerns effectively while maximizing the positive impact of this technology.

\clearpage
{
    \small
    \bibliographystyle{ieeenat_fullname}
    \bibliography{main}
}

\newpage
\clearpage
\maketitlesupplementary

\section{Gaussianity preservation of our noise warping algorithm}

In this section, we discuss our noise warping algorithm, providing a formal proof of its Gaussianity preservation properties. We also present an illustrative example that demonstrates how noise that undergoes expansion and subsequent contraction returns to its original state, showcasing how our noise warping algorithm maintains the underlying Gaussian distribution throughout the warping process.

\begin{proof}
    For each $(x,y) \in V$, $R(x,y)$ is a collection of upsampled noise $X_i$, where
    \begin{align*}
    \bE[X_i] &= \bE[\frac{q(x,y)}{d}] + \bE[\frac{1}{\sqrt{d}}(Z_i - \frac{S}{d})] = 0  \\
        \Var(X_i) &= \Var(\frac{q(x,y)}{d}) + \Var(\frac{1}{\sqrt{d}}(Z_i - \frac{S}{d})) \\
        &= \frac{1}{d^2} + \frac{1}{d} \Var(\frac{d-1}{d}Z_i - \sum_{j\neq i} \frac{Z_j}{d}) \\
        &= \frac{1}{d^2} + \frac{1}{d} \frac{(d-1)^2 + (d-1)}{d^2} = \frac{1}{d},
    \end{align*}
    where we used the fact that $q(x,y)$ and $Z_i$'s are i.i.d. standard Gaussians.
    Since $X_i$ is constructed as a weighted sum of Gaussians, itself is also a Gaussian.
    Moreover, for $i \neq j$, we compute
    \begin{align*}
        &\Cov(X_i, X_j) \\
        =&\Cov(\frac{q(x,y)}{d} + \frac{1}{\sqrt{d}}(Z_i - \frac{S}{d}), \frac{q(x,y)}{d} + \frac{1}{\sqrt{d}}(Z_j - \frac{S}{d})) \\
        =& \frac{1}{d^2} + \frac{1}{d}\bE[(Z_i - \frac{S}{d})(Z_j - \frac{S}{d})] \\
        =& \frac{1}{d^2} + \frac{1}{d}(0 - 2 \frac{\bE[Z_i S]}{d} + \frac{\bE[S^2]}{d^2}) \\
        =& \frac{1}{d^2} + \frac{1}{d}(-\frac{2}{d} + \frac{1}{d}) = 0.
    \end{align*}
    Hence all $X_i$'s are independent.

    For each $(x',y') \in V'$, if $\deg_G((x',y')) = 0$, then $q'(x',y')$ is sampled as an independent standard Gaussian. 
    Otherwise, the output noise pixel $q'(x',y')$ is built as a weighted sum of $R(x,y)\text{.pop}()$ for each edge $((x,y), (x',y'))\in E$, where $R(x,y)\text{.pop}()$ is an independent Gaussian of mean 0 and variance $\frac{1}{\deg_G((x,y))}$.
    Hence $q'(x',y')$ is also a Gaussian with mean 0.
    The variable $s$ after executing the inner for loop thus represents the variance of $q'(x',y')$, so the renormalization at the end brings $q'(x',y')$ back to a standard Gaussian.
    Since the composing $X_i$'s are independent, the resulting noise $q'$ should also have an independent Gaussian in each pixel.
\end{proof}

\begin{example}[Exact recovery of \expansion-\contraction]
Consider the following evolution of noise across three frames with forward flows $f_{i\to j}$ going from frame $i$ to frame $j$ with $i + 1 = j$ (and backward flow if $i -1 = j$).
Suppose at frame $1$, a pixel $v \in D$ with density $1$ has noise $q$. Suppose further that $v'_a$ is a pixel at frame $2$ such that $f_{1\to 2}^{-1}(v'_a) = \{v\}$, and $v'_b \in D$ is the only pixel at frame $2$ such that $f_{1\to 2}^{-1}(v_b') = \varnothing$ and $f_{2\to 1}(v'_b) = v$.
This represents the scenario where $v$ is expanded into two pixels $v'_a,v'_b$.
Then \cref{alg:main} with forward flow $f_{1\to 2}$ and backward flow $f_{2 \to 1}$ will result in $v'_a$ having density $1/2$ and noise $\frac{q}{2} + \frac{1}{\sqrt{2}}(\frac{Z_a-Z_b}{2})$, and $v_b'$ having density $1/2$ and noise $\frac{q}{2} + \frac{1}{\sqrt{2}}(\frac{Z_b-Z_a}{2})$, where $Z_a$ and $Z_b$ are i.i.d. standard Gaussians.
Now, from frame $2$ to frame $3$, suppose there exists a pixel $v''$ such that $f_{2\to 3}^{-1}(v'') = \{v'_a,v'_b\}$, i.e., they both $v'_a$ and $v'_b$ contract to $v''$, and that $f_{3\to2}(D) \cap \{v'_a,v'_b\} = \varnothing$.
Then \cref{alg:main} with forward flow $f_{2\to 3}$ and backward flow $f_{3\to 2}$ will result in $v''$ having density $1$ and noise $q$, hence deterministically recovering the noise and density of $v$ in frame 0.

\end{example}

\section{Qualitative results of training-free image diffusion based video editing}

\begin{figure*}
    \centering
    \includegraphics[width=1\linewidth]{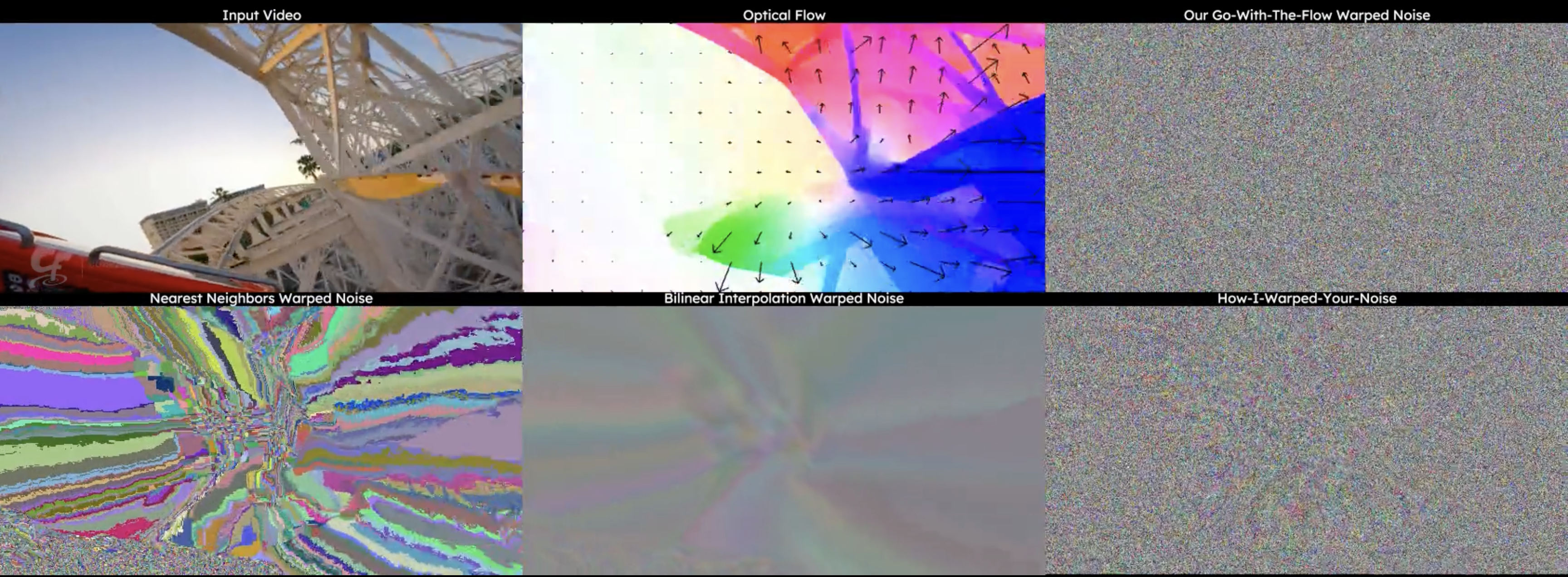}
    \caption{A direct visualization of the noise produced by our noise warping algorithm, HIWYN~\cite{chang2024warped}, bilinear, and nearest neighbor interpolations. The forward movement in this long roller-coaster video forces the noise to expand significantly. Early in the video, the HIWYN baseline produces visibly non-Gaussian results. See the full video on our \href{https://eyeline-research.github.io/Go-with-the-Flow/}{webpage}.}
    \label{fig:supp_warped_noise_flow_vis}
\end{figure*}

\begin{figure}
    \centering
    \includegraphics[width=1\linewidth]{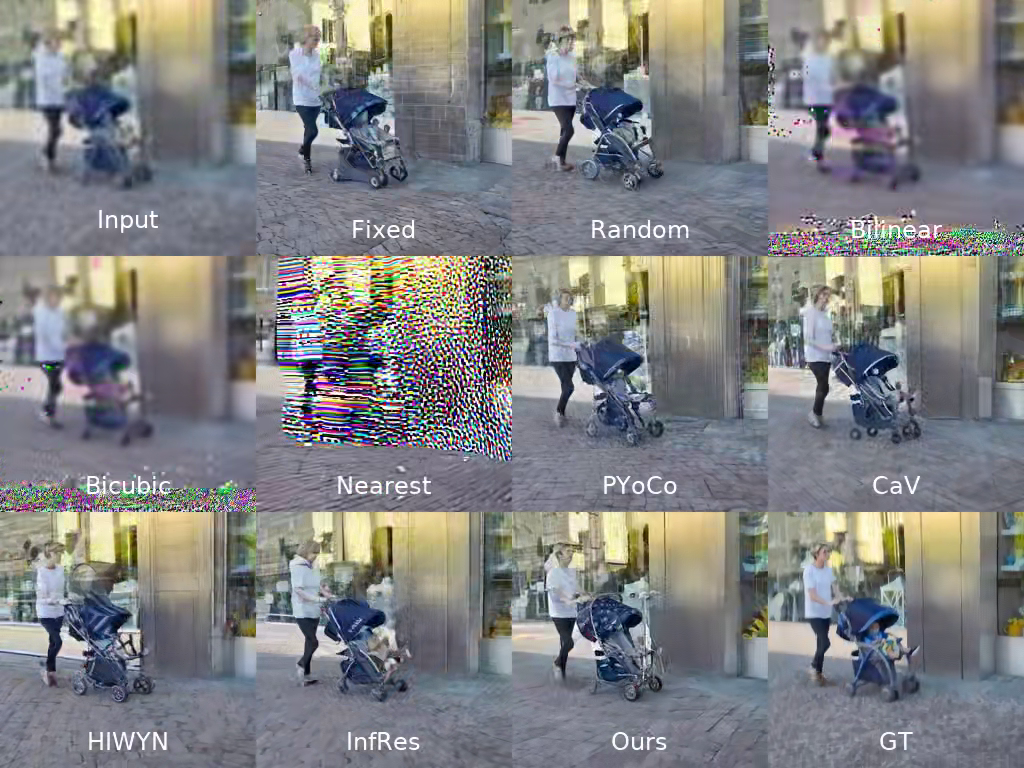}
    \caption{Using different noise warping algorithms on DeepFloyd~IF for video super-resolution on the DAVIS dataset.}
    \label{fig:supp_davis_deepfloyd}
\end{figure}

\begin{figure}
    \centering
    \includegraphics[width=1\linewidth]{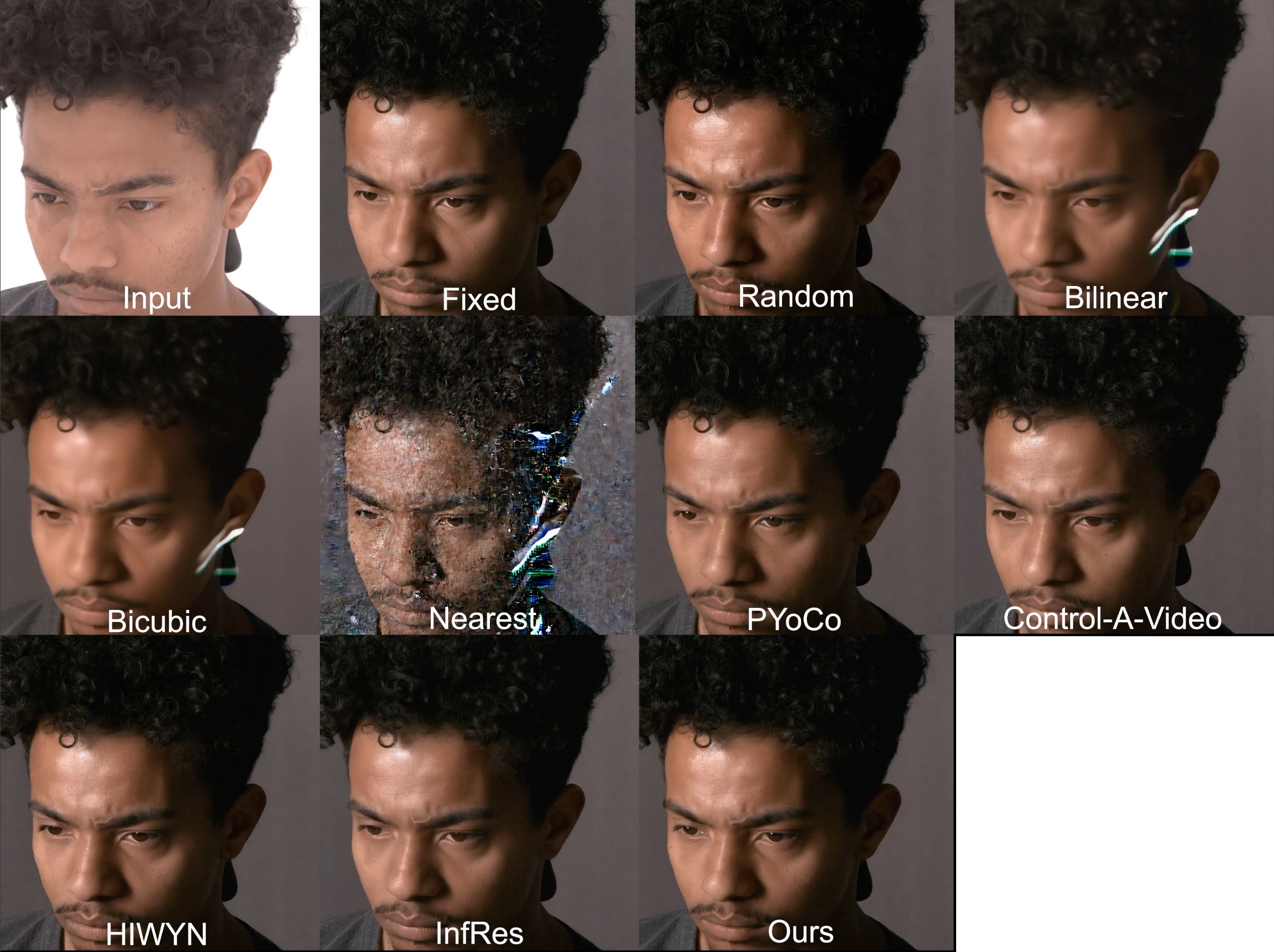}
    \caption{Using different noise warping algorithms on DifFRelight for portrait video relighting. }
    \label{fig:supp_diffrelight_noisewarp}
\end{figure}

Noise warping methods that do not preserve Gaussianity degrade per-frame performance, as originally pointed out in~\cite{chang2024warped}. For example, using nearest neighbor and bilinear interpolation destroys the Gaussianity (see \cref{fig:supp_warped_noise_flow_vis}) and consequently deteriorates the per-frame performance on pre-trained image-to-image diffusion models (see \cref{fig:supp_davis_deepfloyd} and \cref{fig:supp_diffrelight_noisewarp}).

\section{The advantage of noise warping}

By using noise warping as a condition for motion, we effectively discard all structural information from our input video that cannot be inferred from motion alone. This can be advantageous, as demonstrated in \cref{fig:supp_windmill}. MotionClone does not use optical flow to guide the video trajectory, instead relying on manipulating activations within the diffusion model. As a result, the windmill gains an extra set of arms, whereas our method, which relies solely on motion information from optical flow via warped noise, does not introduce such artifacts.

\section{Comparison to the video diffusion base model without finetuning}

\begin{figure*}
    \centering
    \includegraphics[width=0.7\linewidth]{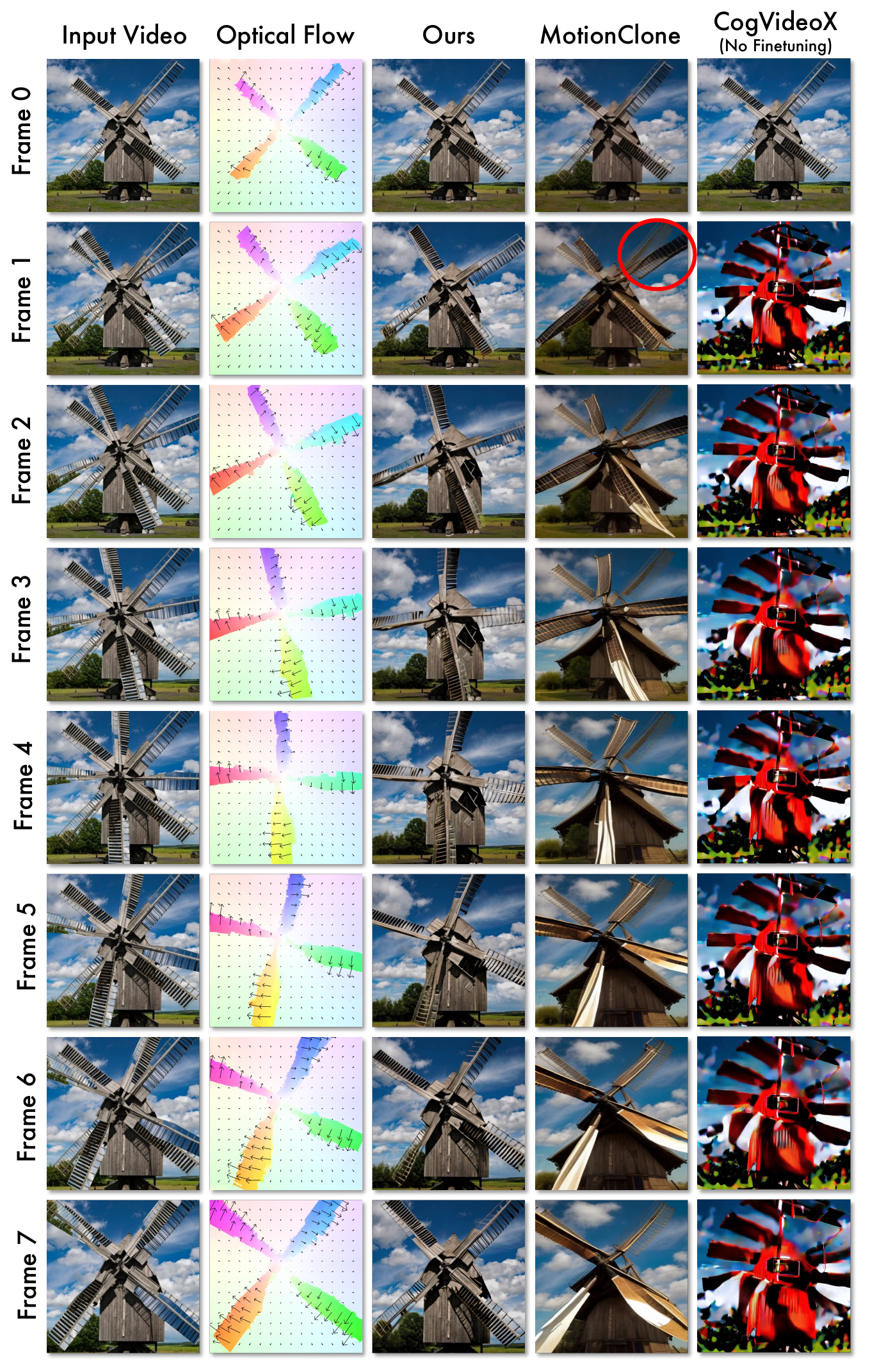} 
    \caption{
    We show a \cutndrag~animation of a windmill rotating clockwise, next to the derived optical flow, our outputs, a baseline and an ablation. \textbf{Note} that the input video column appears to have two sets of panels because it's being cut and dragged over itself to create rotational motion. \textbf{When using noise warping is better}: Per-frame structural information can poison the result of MotionClone, giving the windmill an extra set of arms - whereas ours only receives motion information from optical flow alone via warped noise (there are no double-windmills in the optical flow patterns). \textbf{Ablation in rightmost column}: warped noise with $\deglevel=.5$ on the CogVideoX base model before we fine-tune it. Because warped noise is statisically distinct from the pure Gaussian noise CogVideoX was trained on, without fine-tuning it can result in visual artifacts. Note how although the per-frame quality suffers here, it still picks up on motion queues from the warped noise (the camera zooms into the windmill).}
    \label{fig:supp_windmill}
\end{figure*}

Interestingly, video diffusion models respond to noise warping even without training. In \cref{fig:supp_windmill} the rightmost column, even though the per-frame quality suffers, the flow of the output video still roughly follows the flow of the warped noise. However, because warped noise is statisically distinct from the pure Gaussian noise CogVideoX was trained on, without fine-tuning it can result in visual artifacts.

\section{User study settings and statistics}
\label{sec:supp_user_study}

\begin{figure*}
    \centering
    \begin{subfigure}{.48\linewidth}
    \includegraphics[width=\linewidth]{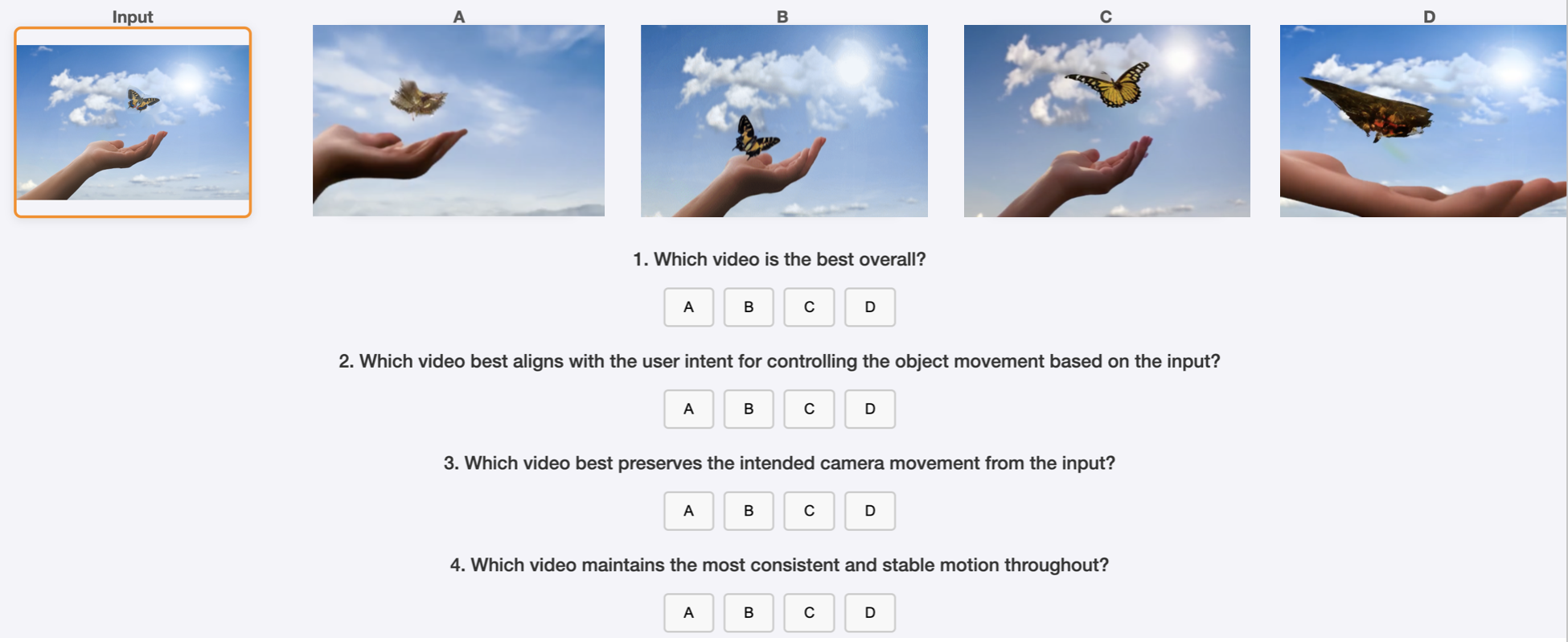}
    \subcaption{User study interface and questions for local object motion control, corresponding to ~\cref{fig:comparisons_video_diffusion_object_motions} in the main paper.}
    \end{subfigure}
    \hfill
    \begin{subfigure}{.48\linewidth}
    \includegraphics[width=\linewidth]{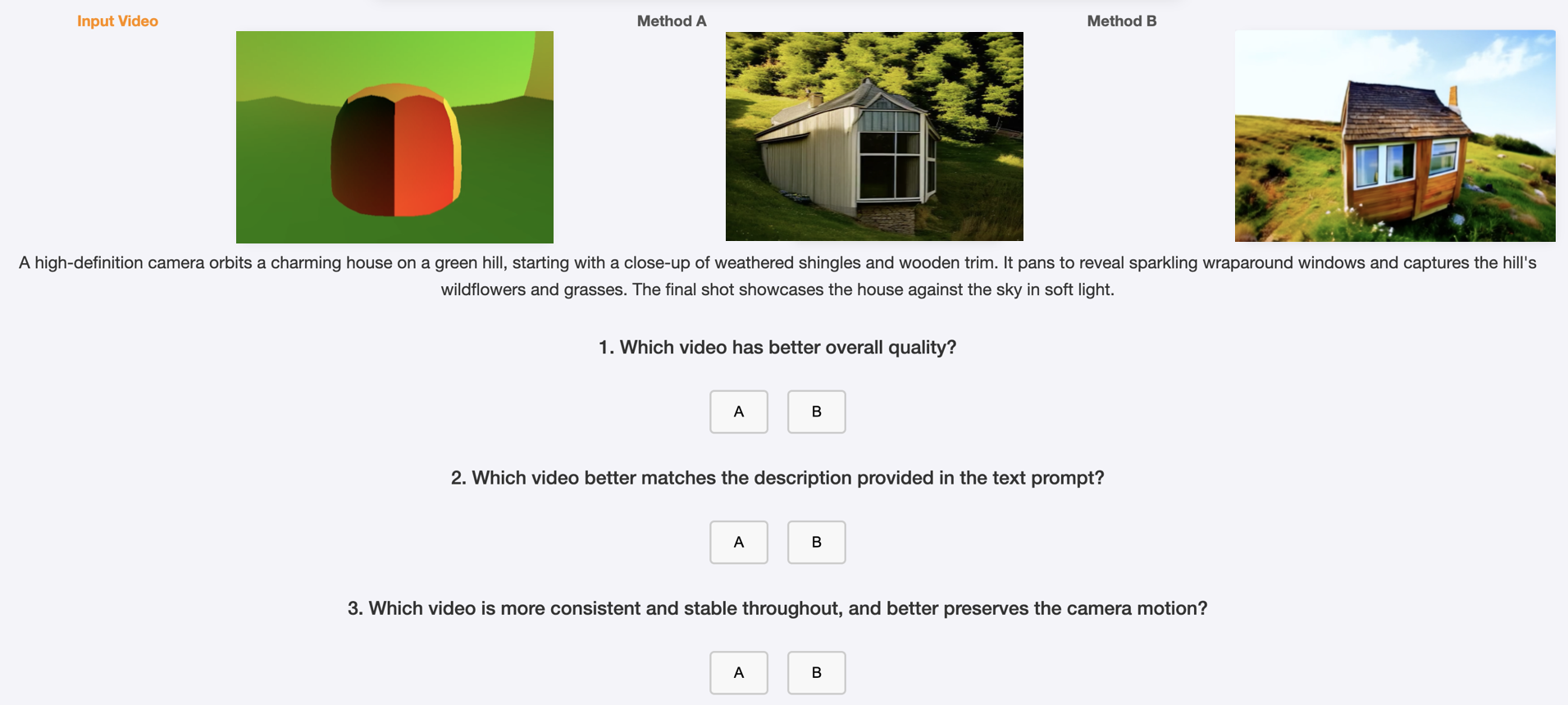}
    \subcaption{User study interface and questions for turnable camera movement video generation, corresponding to ~\cref{fig:comparisons_video_diffusion_turning_object} in the main paper.}
    \end{subfigure}
    \\[12pt]
    \begin{subfigure}{0.48\linewidth}
    \centering
    \includegraphics[width=.7\linewidth]{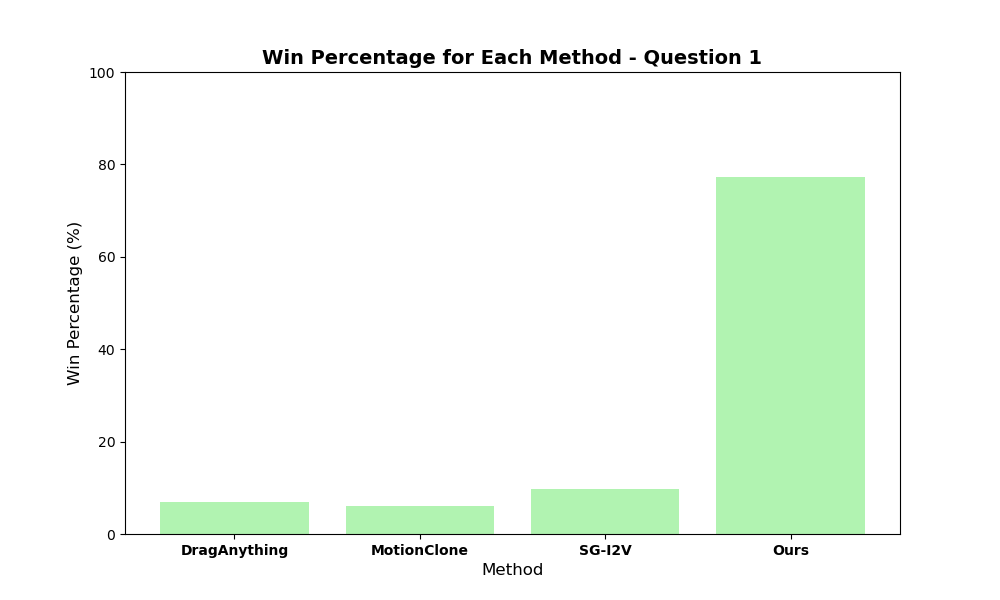}
    \subcaption{User study statistics for local object motion control on the first question ``\textit{Which video is the best overall?}''}
    \end{subfigure}
    \hfill
    \begin{subfigure}{0.48\linewidth}
    \centering
    \includegraphics[width=.7\linewidth]{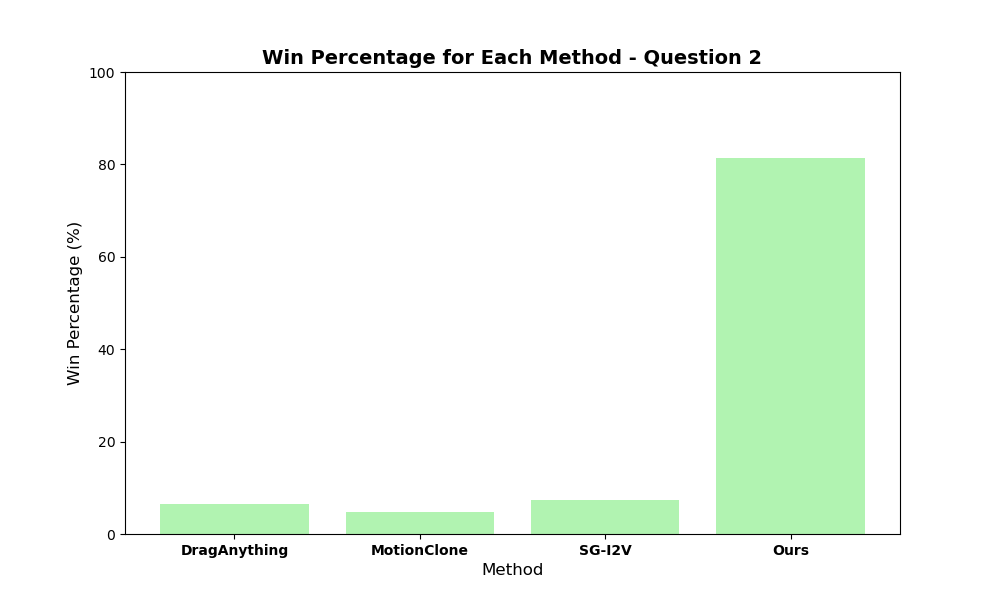}
    \subcaption{User study statistics for local object motion control on the second question ``\textit{Which video best aligns with the user intent for controlling the object movement based on the input?}''}
    \end{subfigure}
    \\[12pt]
    \begin{subfigure}{0.48\linewidth}
    \centering
    \includegraphics[width=.7\linewidth]{fig/user_study_1_statistics_1.png}
    \subcaption{User study statistics for local object motion control on the third question ``\textit{Which video best preserves the intended camera movement from the input?}''}
    \end{subfigure}
    \hfill
    \begin{subfigure}{0.48\linewidth}
    \centering
    \includegraphics[width=.7\linewidth]{fig/user_study_1_statistics_1.png}
    \subcaption{User study statistics for local object motion control on the fourth question ``\textit{Which video maintains the most consistent and stable motion throughout?}''}
    \end{subfigure}
    \\[12pt]
    \begin{subfigure}{0.48\linewidth}
    \centering
    \includegraphics[width=0.7\linewidth]{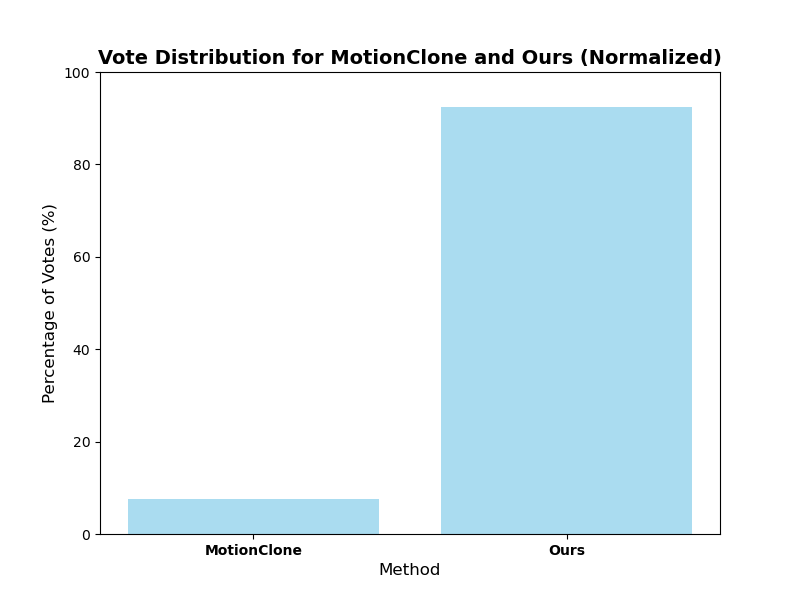}
    \subcaption{User study statistics for motion transfer on the first question ``\textit{Which video has better overall quality?}''}
    \end{subfigure}
    \caption{User study questionnaires screenshots and statistics. For all the questions of both applications, our method (the rightmost bar plot) significantly wins the most user preferences.}
    \label{fig:supp_user_study_screenshots_statistics}
\end{figure*}

\cref{fig:supp_user_study_screenshots_statistics} presents our user study questionnaires and statistics for two applications: (1) local object motion control, and (2) turnable camera movement video generation. Our questions focus on users' overall subjective preference, controllability, and temporal consistency.

\section{Model Agnostic}

\begin{figure*}
    \centering
    \includegraphics[width=1\linewidth]{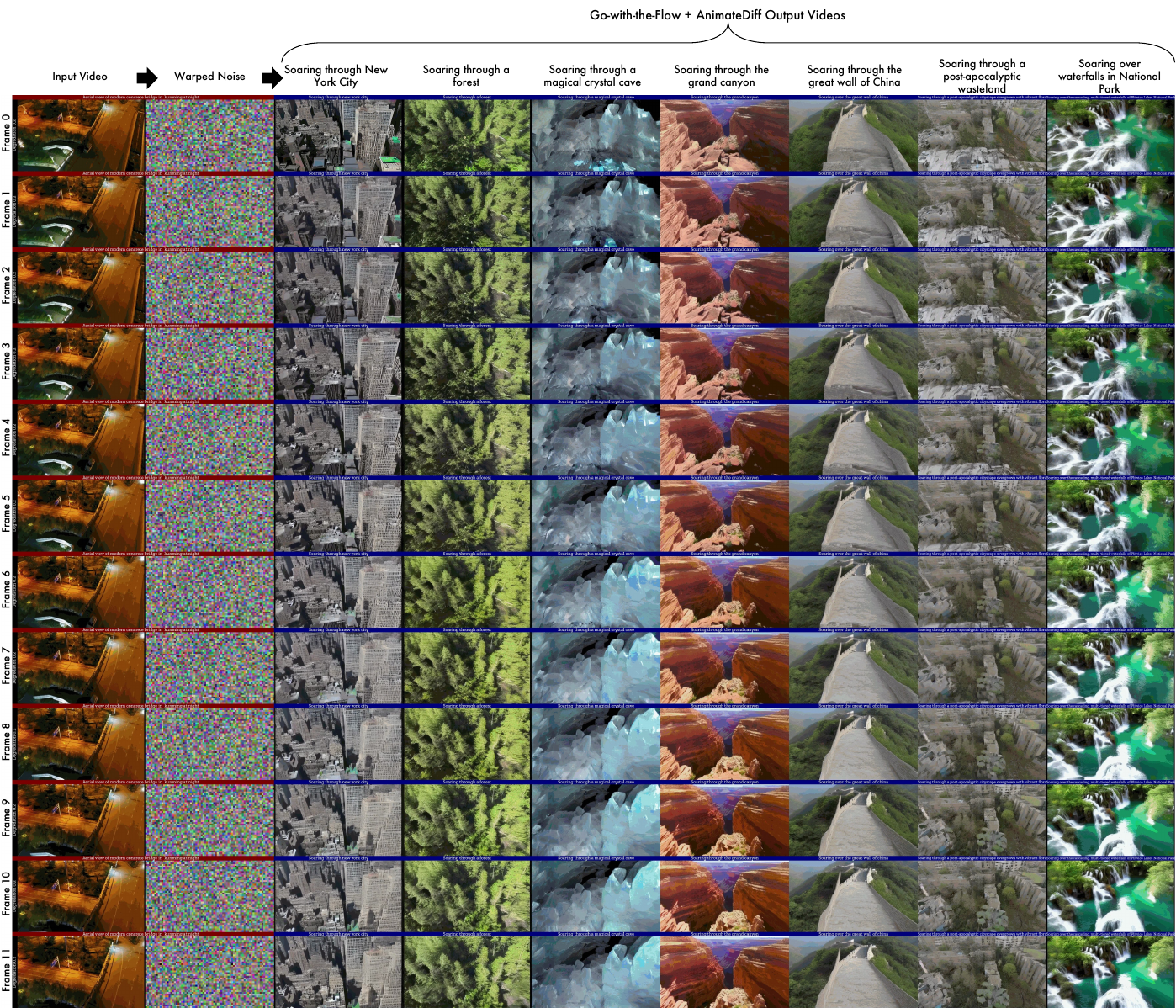}
    \caption{Fine-tuning AnimateDiff with our warped noise flow. We used Go-with-the-Flow to fine-tune AnimateDiff T2V, and display the results above. The input video is on the left, and from that video we derive warped noise which is used to initialize AnimateDiff on the columns to its right with different text prompts.}
    \label{fig:supp_animatediff_grid}
\end{figure*}

Our method is data- and model-agnostic. It can be used to add motion control to arbitrary video diffusion models by only processing the noise sampling during fine-tuning. For example, it also works with AnimateDiff \cite{guo2024animatediff} fine-tuned on the WebVid dataset~\cite{Bain21} (the weights for this model on our \href{https://github.com/Eyeline-Research/Go-with-the-Flow}{GitHub} page). See its qualitative results in \cref{fig:supp_animatediff_grid}. Since release, the community has also trained a version of Go-with-the-Flow on HunyuanVideo (linked on our \href{https://github.com/Eyeline-Research/Go-with-the-Flow}{GitHub} page). Therefore, our method will generalize to future more advanced video diffusion base model.

\section{Pseudo code}

See \cref{listing:supp_algo_pseudo_code} for our noise warping pseudo code. See our source code and model checkpoints on \href{https://github.com/GoWithTheFlowPaper/gowiththeflowpaper.github.io}{GitHub}.

\begin{figure*}
\begin{lstlisting}
def warp_noise(prev_frame, cur_frame, prev_noise, prev_weight):

    height, width, _ = prev_frame.shape

    flow = optical_flow(prev_frame, cur_frame) # Agnostic to the optical flow algorithm
    backwards_flow = -flow # A cheap approximation of optical_flow(cur_frame, prev_frame)

    expansion_noise    = zeros(height, width)
    contraction_noise  = prev_noise.copy()

    expansion_mask     = ones (height, width, type=bool)
    contraction_mask   = zeros(height, width, type=bool)

    for x in range(width): for y in range(height):
        dx, dy = flow[x,y]
        if 0 <= x+dx <= width-1 and 0 <= y+dy <= height-1:
            # This particle stays in bounds
            expansion_mask  [x+dx, y+dx] = False
            contraction_mask[x   , y   ] = True  # Contraction mask is True where 

    for x in range(width): for y in range(height):
        if expansion_mask[x, y]:
            dx, dy = backwards_flow[x,y]
            expansion_noise [x, y] = prev_noise[x+dx, y+dy]

    # We've decided which source pixels are involved in contraction and expansion now
    contraction_noise &= contraction_mask
    expansion_noise, contraction_noise, cur_weight = jointly_regaussianize_and_rebalance_weights(
        expansion_noise, contraction_noise, prev_weight
    ) # Regaussianize all noise values here, and divide the weights by the number of pixels in each bin

    contraction_weight = zeros(height, width)
    for x in range(width): for y in range(height):
        if contraction_mask[x, y]:
            # Contraction treats the noise pixels as particles, each moving from the source to the
            # destination with this flow
            dx, dy = flow[x,y]
            # Contraction is a weighted sum of source pixels to a destination pixel
            pixel_weight = cur_weight[x, y]
            # Sum all the source noise pixels that contract to the same destination
            contraction_noise [x+dx, y+dy] += prev_noise[x, y] * pixel_weight
            # When we multiply a noise pixel by a weight, the variance changes by that weight squared
            contraction_weight[x+dx, y+dy] += pixel_weight ** 2 
    contraction_noise /= sqrt(contraction_weight) # Adjust the variance of the summed contracted noise

    # Mixing contraction and expansion noises with their respective masks
    cur_noise = contraction_noise & contraction_mask + expansion_noise & expansion_mask

    return cur_noise, cur_weight
\end{lstlisting}
\caption{Our noise warping pseudo code.}
\label{listing:supp_algo_pseudo_code}
\end{figure*}

\end{document}